\documentclass[journal,twoside,final,letterpaper,twocolumn]{IEEEtran}

\usepackage{graphicx,amssymb,amsbsy,subfigure,amsmath,amsfonts,multirow,color}
\usepackage{epsfig}
\usepackage{algorithmic}
\usepackage{algorithm}
\usepackage[mathscr]{eucal}
\usepackage{epstopdf}
\usepackage{url}
\usepackage{amsmath}
\usepackage[table,xcdraw]{xcolor}

\graphicspath{{Figure/}}

\begin{document}
\title{Conditional Random Field and Deep Feature Learning for Hyperspectral Image Segmentation}
%
\author{Fahim Irfan Alam, Jun Zhou, \IEEEmembership{Senior Member,~IEEE}, Alan Wee-Chung Liew, \IEEEmembership{Senior Member,~IEEE}, Xiuping Jia, \IEEEmembership{Senior Member,~IEEE}, Jocelyn Chanussot, \IEEEmembership{Fellow,~IEEE}, Yongsheng Gao, \IEEEmembership{Senior Member,~IEEE}
\thanks{F. Alam, J. Zhou, A. Liew and Y. Gao are with the Institute of Integrated and Intelligent Systems, Griffith University, Nathan, Australia. Corresponding author: Jun Zhou (jun.zhou@griffith.edu.au).}
\thanks{X. Jia is with the School of Engineering and Information Technology, University of New South Wales at Canberra, ACT, Australia.}
\thanks{J. Chanussot is with Laboratoire Grenoblois de l'Image, de la Parole, du Signal et de l'Automatique (GIPSA-Lab), Grenoble Institute of Technology,
38402 Saint Martin d'Heres Cedex, France, and also with the Faculty of Electrical and Computer Engineering, University of Iceland, 107 Reykjav\'{i}k,
Iceland.}
}

\maketitle

%
%
\begin{abstract}
Image segmentation is considered to be one of the critical tasks in hyperspectral remote sensing image processing. Recently, convolutional neural network (CNN) has established itself as a powerful model in segmentation and classification by demonstrating excellent performances. The use of a graphical model such as a conditional random field (CRF) contributes further in capturing contextual information and thus improving the segmentation performance. In this paper, we propose a method to segment hyperspectral images by considering both spectral and spatial information via a combined framework consisting of CNN and CRF. We use multiple spectral cubes to learn deep features using CNN, and then formulate deep CRF with CNN-based unary and pairwise potential functions to effectively extract the semantic correlations between patches consisting of three-dimensional data cubes. Furthermore, we introduce a deep deconvolution network that improves the segmentation masks. We also introduced a new dataset and experimented our proposed method on it along with several widely adopted benchmark datasets to evaluate the effectiveness of our method. By comparing our results with those from several state-of-the-art models, we show the promising potential of our method.
\end{abstract}
\section{Introduction}
Hyperspectral imaging technology acquires and analyses images in contiguous spectral bands over a given spectral range~\cite{HSI}. It enables more accurate and detailed spectral information extraction than is possible with other types of remotely sensed data. This capability has greatly benefited the identification and classification of spectrally similar materials. Along with spectral information, the spatial relationships among various spectral responses in a neighborhood can be explored, which allows development of spectral-spatial models for accurate image segmentation and classification. Thanks to these advantages, hyperspectral imaging has become a valuable tool for a wide range of remote sensing applications in agriculture, mineralogy, surveillance and environmental sciences~\cite{Remote}.

The research in hyperspectral image segmentation is faced with several challenges. The unbalance between large dimensionality of spectral bands and insufficient training samples impose a major restriction on segmentation performance. Segmentation algorithms exploiting only the spectral information fail to capture significant spatial variability of spectral signatures for the same class of targets~\cite{kernel} and therefore result in unsatisfactory performance. With such critical issues unsolved, hyperspectral segmentation faces a major drawback in practical usage. Several strategies can be adopted to overcome these problems. An effective solution is to design algorithms using both spectral and spatial information, which provides more discriminative information regarding object shape, size, material, and other important features~\cite{Spec-Spatial}.

Spectral-spatial segmentation methods can be divided into two categories. The first category uses the spectral and spatial information separately in which the spatial information is perceived in advance by the use of spatial filters~\cite{attribute}. After that, these spatial features are added to the spectral data at each pixel. Then dimensionality reduction methods can be used before the final classification and segmentation. The spatial information can also be used to refine the initial pixelwise classification result as a post-processing step, e.g., via mean shift~\cite{meanshift} or Markov random field~\cite{SVM-MRF}, which is a very common strategy in image segmentation~\cite{MRF2,MRF-seg}. The second category combines spectral and spatial information for segmentation. Li \emph{et al.} proposed to integrates the spectral and spatial information in a Bayesian framework, and then use either supervised~\cite{Li} or semi-supervised algorithm~\cite{ML} to perform segmentation. Yuan \emph{et al.}~\cite{spectraltexture} combined spectral and texture information where linear filters were used to supply enhanced spatial patterns. Gills \emph{et al.}~\cite{graph} modelled a hyperspectral image as a weighted graph where the edge weights were given by a weighted combination of the spectral and spatial information between the nodes. Besides feature extraction step, traditional image classification or segmentation methods, such as watershed algorithm~\cite{Watershed} and minimum spanning forest~\cite{MSF}, have been adopted to perform joint spectral-spatial processing of hyperspectral image. Since hyperspectral data are normally represented in three-dimensional cubes, the second category of methods can result in a large number of features containing discriminative information which are effective for better segmentation performance.

Recent advances in training multilayer neural networks have contributed much in a wide variety of machine learning problems including classification or regression tasks. The ``deep'' architecture can extract more abstract and invariant features of data, and thus have the ability to produce higher classification accuracy than the traditional classifiers~\cite{DL}. It has also demonstrated its success in classifying spectral-spatial features~\cite{Jia,DeepCNN}. Amongst various deep learning models, convolutional neural network (CNN) has been widely used for pixel-level labeling problems. With this model, good representation of features can be learned, which allows performing an end-to-end labeling task~\cite{FCN}. Recently, this model was adopted by Chen \emph{et al.}~\cite{yushi} for feature extraction and classification of hyperspectral images based on three-dimensional spectral cubes across all the bands that combines both spectral and spatial information. Similar works have been proposed to extract spectral-spatial features from pixel or pixel-pairs using deep CNN~\cite{shiqi,blde-cnn,pixel-pair}.

\textcolor[rgb]{0.00,0.00,1.00}{Because CNN can effectively discover spatial structures among the neighboring patches of the input data, the resulting classification maps generally appear smoother in spite of not modeling the neighborhood dependencies directly. However, the possibility of reaching local minima during training of CNN and the presence of noise in the input images may create holes or isolated regions in the classification map. Compared with other machine learning methods, CNN is generally limited by the absence of shape and edge constraints. As a result, the final segmentation appears rough on edges. Moreover, in hyperspectral image remote sensing, cloud shadows and topography cause variations in contrast, which often generates incorrect classes in images. The presence of cloud also may hide regions or decreases the contrast of regions. Due to these reasons, CNN sometimes recognizes only parts of the regions properly~\cite{CNN-prob}.}

\textcolor[rgb]{0.00,0.00,1.00}{In these circumstances, a further step of segmentation produce much refined output. To this end, combining probabilistic graphical models such as Markov Random Field (MRF) and Conditional Random Field (CRF) with CNN brings significant improvements by explicitly modelling the dependencies between regions. CRFs have been used to model important geometric characteristics such as shape, region connectivity, contextual information between regions and so on. For these reasons, there has been a recent trend on exploring the integration of CNN and CRF methods~\cite{CNN-CRF2,CNN-CRF3,CNN-CRF4,CNN-CRF5}. For example, Liu et. al~\cite{CRF-CNN} used CRF to refine the segmentation outputs as a post-processing step. However, the CRF is entirely disconnected from the training of the deep network. Instead of using a disconnected approach, Chen et al.~\cite{deconvo} proposed a fully connected Gaussian CRF model where the respective unary potentials were supplied by a CNN.  Since CRF can directly model spatial structures, if it can be formulated in a deep modeling approach, the trained model will integrate the advantages of both CNN and CRF in modeling the spatial relationships in the data. Based on this combination, CRFs can better fine-tune model features by using the incredible power of CNNs.}

Some recent methods combine CNNs and CRFs for semantic segmentation. For example, Zheng \emph{et al.}~\cite{CRF-RNN} formulated a dense CRF with Gaussian pairwise potentials as a Recurrent Neural Network (RNN) that was used to refine the upsampled low-resolution prediction by a traditional CNN. We argue that this stage of refinement process can be further improved by applying more advanced refinement methods, such as training a deconvolution network~\cite{FCN}. Deconvolution network~\cite{deconvo3} was employed to visualize activated features in a trained CNN and update network architecture for performance enhancement. It is plausible to use it in a segmentation framework to further improve the output map obtained from earlier steps.

In computer vision applications, it is usually common to train a deep network with large amount of samples. This, however, is a very challenging task in hyperspectral remote sensing applications since the number of training samples is limited. Without abundant training samples, a deep network faces the problem of ``overfitting'' which means the representation capability may not be sufficient to perform well on test data. It is therefore very important to increase the size of the training samples in order to handle this overfitting issue.

\begin{figure*}
\centering
\includegraphics[scale=0.28]{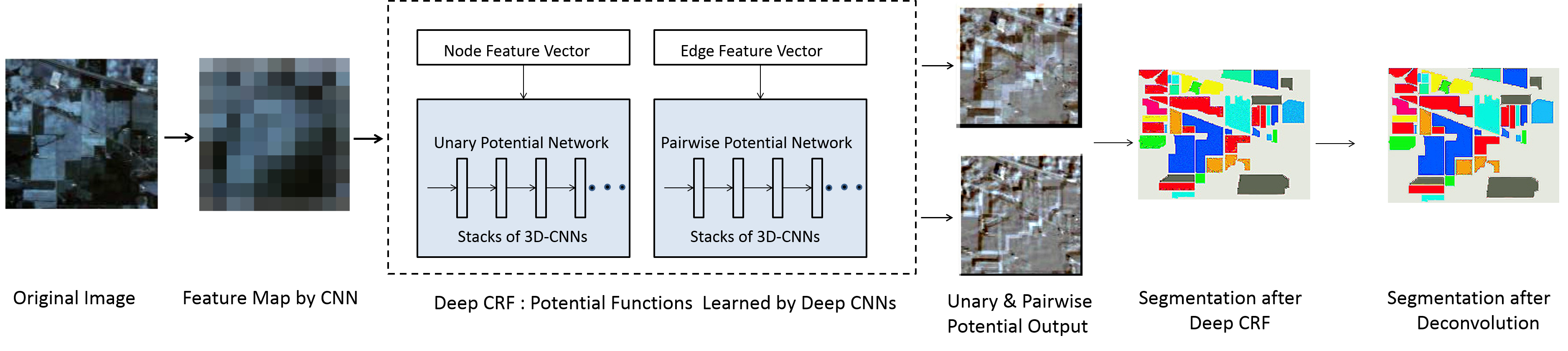}
\caption{Proposed Architecture of our method. The original image is converted into a feature map by a 3D CNN which is later used to train the proposed deep CRF in order to generate unary and pairwise potentials. The output of deep CRF with deconvolution is later used to perform the final prediction.}
\label{fig:framework}
\end{figure*}

In our method, we treat hyperspectral images as spectral cubes consisting of the image spanning over a few spectral bands instead of all the bands across the spectra as in~\cite{yushi}. Such smaller-sized yet a large number of spectral cubes will be able to provide more accurate local spectral-spatial structure description of the data. Our framework, ``3DCNN-CRF'' (illustrated in Fig.~\ref{fig:framework}) starts with generating a coarse feature map obtained by applying a CNN over the spectral cubes which later constructs a CRF graph. We then calculate the unary and pairwise potentials of the CRF by extending a CNN-based deep CRF architecture~\cite{piecewise2} to cope with both spectral and spatial information along the entire spectral channels. Finally, a mean-field inference algorithm~\cite{mean} was used to perform the final segmentation.
The main contributions of this paper are as follows:
\begin{itemize}
\item 3DCNN is performed on spectral cubes containing a small number of bands, which results in a more effective spectral-spatial structure representation of the hyperspectral data for the initial classification.
\item CNN-based general pairwise potential functions in CRFs are extended to explicitly model the spatial relations among the spectral cubes to perform the segmentation.
\item 3DCNN-CRF also learns a deep deconvolution network during CRF pairwise potential training which is composed of deconvolution, unpooling, and rectified linear unit (ReLU) layers. Learning deep deconvolution networks improved the segmentation performance to a considerable extent which is very relevant and practical in the context of semantic segmentation.
 \item The size of the training samples is augmented by generating virtual samples from real ones in different band groups, which produces different yet useful estimation of spectral-spatial characteristics of the new samples.
\item A new hyperspectral dataset is created and the image regions containing relevant classes are manually labelled. 3DCNN-CRF is evaluated on this dataset and compared with alternative approaches accordingly.
\end{itemize}

\section{3D CNN for Hyperspectral Feature Representation}
Deep learning techniques automatically learn hierarchical features from raw input data. When applied to hyperspectral data, unsupervised deep learning methods such as stacked autoencoder (SAE)~\cite{Autoencoder} and deep belief network (DBN)~\cite{Jia} can extract deep features in a layer-wise training fashion. However, the training samples composed of image patches have to be converted into vectors in order to fit the input layer of these models, which can not retain the original spatial information in the image. Same problem applies to 1D-CNN models when vectorized feature is used as the input. These methods normally extract the spatial and spectral features separately and therefore, they do not fully exploit the important joint spectral-spatial correlation of image data which is very important for segmentation.

In some CNN-based hyperspectral data classification methods~\cite{blde-cnn}, the spatial features are obtained by a 2D-CNN model which exploits the first few principal component (PC) bands of the original hyperspectral data. Yu \emph{et al.}~\cite{shiqi} proposed a CNN architecture which uses a convolutional kernel to extract spectral features along the spectral domain only. To obtain features in the spatial domain, they used normalization layers and a global average pooling (GAP) layer.  On the other hand, 3D-CNN can learn the signal changes in both spatial and spectral dimensions of local spectral cubes. Therefore, it can exploit significant discriminative information for segmentation and form powerful structural characteristics for hyperspectral data.

\textcolor[rgb]{0.00,0.00,1.00}{In this paper, we use 3D CNN to obtain  effective spectral-spatial structure representation of the hyperspectral data for the initial classification. 3D kernels are used for the important convolution operations on the spectral cubes. A common practice is to convolve the 3D kernels with a spatial neighborhood along the entire spectral channels. However, in our method, the original image, which has \textit{B} bands, is divided spectrally into several images consisting of neighboring \textit{L} bands where $L \ll B$. 3D convolution filters are applied to each of these different band group images. These groups of bands provide more detailed local spatial-spectral information so as to let different wavelength ranges make distinct contribution to the final classification/segmentation outcome. Repeated convolution operations produce multiple feature maps along the spectral cubes. Let $(x,y)$ define a location in the spatial dimension and $z$ be the band index in the spectral dimension. The value at a position $(x,y,z)$ on the $c^{th}$ feature map is given by~\cite{3D-CNN-2}:}
\begin{equation}
val^{xyz}_{lj_{(c)}}=f(\sum_{i=1}^{m} \sum_{p=0}^{P_l-1}\sum_{q=0}^{Q_l-1}\sum_{r=0}^{R_l-1}k_{lij}^{pqr}val_{(l-1)ij}^{(x+p)(y+q)(z+r)}+b_{lj})
\end{equation}
where $l$ is the current layer that is being considered, $m$ is the number of feature maps in the {$(l-1)$-th} layer (previous layer), $j$ is the current kernel number, $i$ is the current feature map in the $(l-1)$-th layer connected to the feature map of the $l$-th layer, $k_{lij}^{pqr}$ is the $(p,q,r)$-th value of the kernel connected to the $i$-th feature map in the previous layer. $P_l$ and $Q_l$ are the height and the width of the kernel respectively, and $R_l$ is the size of the kernel along the spectral dimension.

Each feature map is treated independently. Therefore, $val^{xyz}_{lij}$ is calculated by convolving a feature map of the previous layer with a kernel of the current layer. In this process, the number of feature maps in the previous layer will be multiplied by the number of kernels in the current layer which will produce as many feature maps as the output of the $l$-th convolution layer. Therefore, 3D convolution can preserve the spectral information of the input data.

\begin{figure*}
\centering
\includegraphics[scale=0.25]{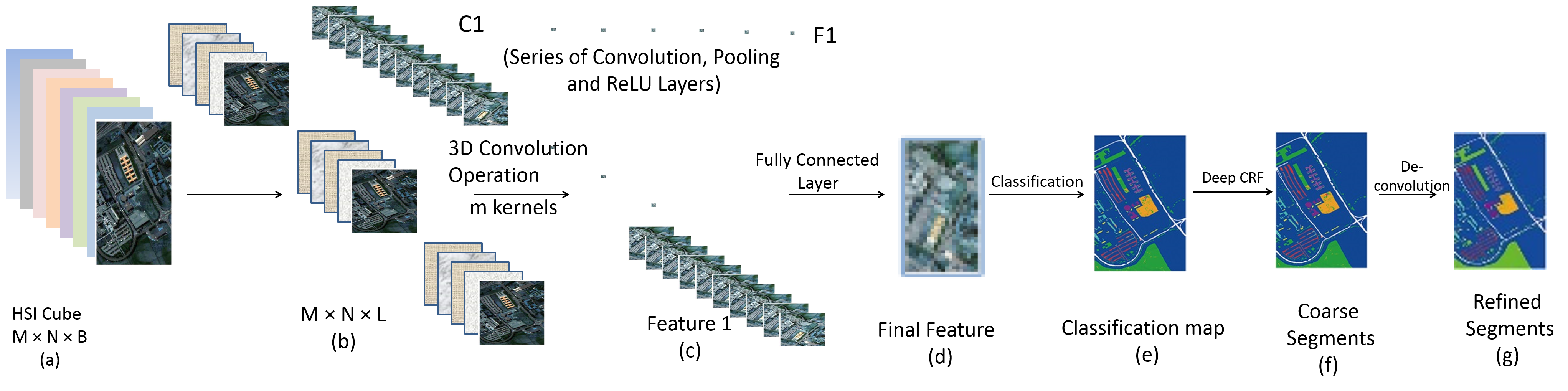}
\caption{Working flow of 3D CNNs. (a) The original Hyperspectral cube with B bands. (b) Different band groups consisting of L bands ($L\ll B$) representing our input. (c) Resulting feature maps after C1 convolution. (d) Resulting final feature map after applying series of convolutional, pooling and activation function layers. (e) Classification map (f) Segmentation map after applying 3D CNN-based deep CRF. (g) Segmentation map after applying deconvolution.}
\label{diagram}
\end{figure*}

After the convolution operations between the kernels and the spectral cubes, the intermediate feature maps pass through the pooling layers and the activation functions. Finally, the feature maps consisting of the data cubes are flattened into a feature vector and feed into a fully-connected layer which extracts the final learned deep spectral-spatial features. The output is later fed to a simple linear classifier such as softmax function to generate the required classification result. The entire network is trained using the standard back propagation algorithm. Fig.~\ref{diagram} illustrates the working flow of the 3D CNNs used in our proposed framework.
 \textcolor[rgb]{0.00,0.00,1.00}{After the CNN training, the learned parameters $\theta_\lambda$ contain information distinct to each band group along the spectral channel $\lambda \in B$. Such representation is very useful for deep learning framework as the model will be able to receive as much information as required to identify interesting patterns within the data. The procedure of obtaining spectral-spatial features by CNN is summarized in Algorithm 1.}

\begin{algorithm}
\caption{\color{blue}3D-CNN Training Algorithm}
\color{blue}
\hspace*{\algorithmicindent} \textbf{Input:} T Input Samples \{$X_1$, $X_2$,\dots,$X_T$\},
T target labels in \{$Y_1$, $Y_2$,\dots,$Y_T$\},
number of BP epochs R

\begin{algorithmic}[1]
\FOR{Each subcube $M \times N \times L$ in $\lambda$}
\WHILE{epoch $r:1 \rightarrow R$}
\WHILE{training sample $i:1 \rightarrow T$}
        \STATE Perform convolution operations using Eq. (1) to generate intermediate feature maps.
        \STATE Compute Soft-max activation $a=\frac{exp(o)}{\sum_kexp(o_k)}$; where $o$ is the output of the final layer of the network and first input to softmax classifier
        \STATE Compute error $E=\textbf{y}_i$-\textbf{a}
        \STATE Back-propagate error to compute gradient $\frac{\delta E}{\delta o_j}$
        \STATE Update network parameter $\theta_\lambda$ using gradient descent $\Delta w_{ij}=-\eta \frac{\delta E}{\delta w_{ij}}$ where $\eta$ is the learning rate
\ENDWHILE
\ENDWHILE
\ENDFOR
\end{algorithmic}
\hspace{\algorithmicindent} \textbf{Output:} Trained CNN parameters $\theta_\lambda$
\end{algorithm}
\textcolor[rgb]{0.00,0.00,1.00}{The resulting 3D features map is used to formulate our proposed deep CRF as explained in section~\ref{deepcrf}.}

\subsection{\textcolor[rgb]{0.00,0.00,1.00}{Addition of Virtual Samples}}\label{virtual-sample}

\textcolor[rgb]{0.0,0.00,1.00}{In many occasions, substantial number of weights in a CNN introduces local minima in the loss function and eventually restricts the classification performance. To overcome this issue, large amount of samples can be used to update weights during the training procedure. Unfortunately, the process of obtaining samples, which normally requires manual labelling, is time consuming and expensive. In remote sensing applications, the number of available training samples is usually limited. This imposes a great deal of challenges to the adoption of a deep network model.}

\textcolor[rgb]{0.00,0.00,1.00}{To address this issue, the size of the training samples can be augmented by virtually sampling from the existing labelled samples. Remote sensing scenes generally cover large areas where pixels belonging to same class in different locations fall under different radiations. We can simulate this lighting condition in order to generate virtual samples by multiplying with a random factor and also adding a gaussian noise. A virtual sample $y_{(\lambda)}$ is therefore generated from two real samples of the same class represented by  $x_{i_{(\lambda)}}$ and $x_{j_{(\lambda)}}$ along the spectral channel $\lambda \in B$ by
\begin{equation}
y_{(\lambda)}=\alpha_ix_{i_{(\lambda)}}+(1-\alpha_i)x_{j_{(\lambda)}}+\beta
\end{equation}
where $\alpha$ indicates the effects of light intensity, which vary under different atmospheric conditions and $\beta$ is the added random Gaussian noise, which may result from the interactions of adjacent pixels.}

\textcolor[rgb]{0.00,0.00,1.00}{$y_{(\lambda)}$ is assigned with the same class label as $\alpha_ix_{i_{(\lambda)}}$ since the hyperspectral characteristics
of the new fused virtual sample shall be sitting between $x_{i_{(\lambda)}}$ and $x_{j_{(\lambda)}}$ which belong to the same class. Moreover, we generate virtual samples within band groups and hence, they give us multiple spectral information of the same sample from different wavelengths so as to further augmenting the training data. We insert these new samples into separate images by replacing the real samples which are used for the fusion. The original image containing the limited number of real samples and the augmented images containing the virtual samples obtained from different wavelengths are used for training the CNN.}

\textcolor[rgb]{0.00,0.00,1.00}{We further augment the training samples by transforming the sample pixels using rotation ($90^{\circ}$, $180^{\circ}$, $270^{\circ}$) and flipping operations. This leads to 7 combinations which significantly increase the amount of training data. Finally, the smaller number of real samples and the augmented virtual samples, generated by both sample fusion and transformation operations, are used together to train the CNNs. During the training of the CNNs, part of the total number of real and virtual samples are used for learning the weights of the neurons and the rest are used for validating and updating the architecture.}

\section{Constructing Deep CRFs For Segmentation}
\label{deepcrf}
After applying 3D CNN into our original data, we obtain initial classification outcome from the final feature map generated by the CNN. \textcolor[rgb]{0.00,0.00,1.00}{In many cases, CNN can effectively discover spatial structures among the neighboring patches of the input data, which generates smooth classification maps generally in spite of not modeling the neighborhood dependencies directly. However, it still encounters several problems such as}
\begin{itemize}
\item \textcolor[rgb]{0.00,0.00,1.00}{There are holes or disconnected regions in the classification map obtained by a CNN due to the occurrence of reaching local minima during the training.}
\item \textcolor[rgb]{0.00,0.00,1.00}{CNN is generally limited by the absence of shape and edge constraints. The final segmentation may appear rough on edges of some regions or objects.}
\item \textcolor[rgb]{0.00,0.00,1.00}{In hyperspectral remote sensing, cloud shadows and topography cause variations in the spectral responses and influence the contrast of regions, which generates incorrect classes in images. As a result, the CNN sometimes recognizes only parts of the regions,
particularly as observed in the Griffith-USGS dataset.}
\end{itemize}

\textcolor[rgb]{0.00,0.00,1.00}{To resolve such critical issues, an additional step of segmentation will contribute greatly in improving the initial classification output and produce much refined segments across the image. We, therefore, propose an end-to-end modelling approach by integrating CRF with CNN in order to utilize the properties of both CNN and CRF to better characterize the spatial contextual dependencies between the regions. We believe that such end-to-end learning approach is very suitable for hyperspectral image analysis as the integrated models will make full use of the spatial relationships among spectral cubes to perform the segmentation.  This is the motivation of our work.}

In this section, we briefly explain the working principles of this deep CRF employed in our framework. The deep CRF model was motivated by the work of Lin \emph{et al.}~\cite{piecewise2} which works on color or grayscale images. We have significantly extended this model to cope with the spectral dimension of the data. CRF is an undirected probabilistic graphical model. It has powerful reasoning ability and is able to train complex features. During the training, CRF makes full use of the spatial contextual information in the process which is very relevant and useful in hyperspectral applications.

In this paper, we propose a deep CRF that will further analyses the output obtained by the 3D CNN described in the previous section. It is important to note that the output provided by the 3D CNN is in the form of 3D feature maps whose individual location is defined by spatial co-ordinates along the spectral domain. We define these spectral-spatial locations as \textit{voxels}. Our proposed deep CRF is capable of modelling these voxel-neighborhoods, therefore making it ideal for processing hyperspectral data. The parameters of the deep CRF used in our method were trained by stacks of CNNs applied on the initial feature map. However, instead of using group of bands, the CNNs used in the deep CRF consider the entire spectral channels together as input to the network since the initial feature map is already a powerful representation of local spectral-spatial features with different wavelength ranges.

The nodes in the CRF graph correspond to each voxel in the feature map along the B bands. The labels of the voxels are given by $l \in Y$. Later, edges are formed between the nodes which construct the pairwise connections between neighboring voxels in the CRF graph by connecting one node to all other neighboring nodes. The CRF model can be expressed as
\begin{equation}
P(l|v_{(d,\lambda)};\theta_\lambda)=\frac{1}{Z(v_{(d,\lambda)})}exp(-E(l,v_{(d,\lambda)};\theta_\lambda))
  \label{CRF-model}
\end{equation}
where the network parameters $\theta$ along different wavelengths $\lambda$ shall be learned. $E(l,v_{(d,\lambda)};\theta_\lambda)$ is the energy function that models how compatible the input voxel $v$ is. $v$ is defined by spatial co-ordinates $d=\{x,y\}$ along the spectral domain $\lambda$ and is with a particular predicted label $l$. $Z(v_{(d,\lambda)})=\sum exp[-E(l,v_{(d,\lambda)};\theta_\lambda)]$ is the partition function. In order to combine more useful contextual information, we should model the relationships between the voxels in the CRF graph. Therefore, the energy function can be expressed as
\begin{equation}
E(l,v_{(d,\lambda)};\theta_\lambda)=\sum_{\substack{p\in M \\ \lambda \in B}} \phi(l_p,v_p;\theta_\lambda)+\sum_{\substack{(p,q) \in N \\ \lambda \in B}}\psi(l_p,l_q,v_p,v_q;\theta_\lambda)
\label{CRF-energy}
\end{equation}
where $M$ is the total number of voxels/nodes and $N$ is the total number of edges between the nodes in the CRF graph. Here $\phi$ is a unary potential function calculated for individual voxels, and $\psi$ is a pairwise potential function which is determined based on the compatibility among adjacent voxels. In our method, we used 4-connectivity to connect each node horizontally and vertically with four neighboring nodes that have the spatial coordinates $(x \pm 1,y)$ or $(x,y \pm 1)$ instead of connecting all other nodes in order to reduce the computational complexity.

\subsection{Unary potential functions}

In our deep CRF model, we apply stack of 3D CNNs and generate feature maps and a fully connected layer to produce the final output of the unary potential at each individual voxel along $\lambda$. The stack of 3D CNNs was applied on the node feature vectors, obtained from the initial feature map, to calculate the unary potential for each individual voxel representing nodes in the CRF graph.

The unary potential function $\phi$ is computed as follows:
\begin{equation}
\phi(l_p,v_p;\theta_\lambda)=-logP(l_p|v_p;\theta_\lambda)
\end{equation}
It is important to note that the network parameters $\theta_\lambda$ are adjusted in the first CNN according to $\lambda$ for different band groups that provide much useful and discriminative information about the data. However, during the deep CRF training, $\theta_\lambda$ are adjusted in the stack of CNNs along the entire spectral channels as they no longer are grouped into groups of bands.
\subsection{Pairwise potential functions}
The pairwise potential functions are calculated by considering the compatibility between the pair of voxels for all possible combinations (in our case, four adjacent voxels). As the first CNN applied to the original image gives us the feature vector for individual voxels in the feature map, therefore, the edge features can be formed by concatenating the feature vectors of two adjacent voxels~\cite{CRF-train}. Stack of 3D CNNs are then applied on the edge feature vectors which eventually gives us the pairwise potential output. The pairwise potential function is expressed as follows:
\begin{equation}
\psi(l_p,l_q,v_p,v_q;\theta_\lambda)=\mu(v_p,v_q)\delta_{p,q,l_p,l_q}(f_p,f_q;\theta_\lambda)
\end{equation}
Here $\mu()$ is the label compatibility function which encodes the possibility of the voxel pair $(v_p, v_q)$ being labeled as $(l_p,l_q)$ by taking the possible combinations of pairs. $\delta_{p,q,l_p,l_q}$ is the output value of the CNNs applied to the pair of nodes that are described by the corresponding feature vectors $f_p$,$f_q$ previously obtained by the initial CNN. $\theta_\lambda$ contains the CNN parameters to be learned for the pairwise potential function along the whole spectral channels $\lambda$.

After the unary and pairwise potentials from the deep CRFs are obtained, future instances of test images can be segmented by performing a CRF inference on our model, for which a mean-field inference algorithm is adopted.

\subsection{Mean-field Inference}
CRF inference is performed on our deep model to obtain the final prediction of a test image. By minimizing the CRF energy, we obtain the most probable label assignment for the voxels. But in practice, due to large number of parameters contained in the CRF energy function for both unary and pairwise potentials, this exact minimization is nearly impossible. To this end, the mean-field approximation algorithm~\cite{mean} is used to calculate the CRF distribution for maximum posterior marginal inference. This is implemented by approximating the CRF distribution $P(v)$ by a simpler distribution $Q(v)$ which is expressed as the product of independent marginal distributions.

In this CRF model, we use two Gaussian kernels that operate in the feature space defined by the intensity of voxel $v$ at coordinates $d=\{x,y\}$ and the spectral band $\lambda$. For the task of segmentation, we use those two-kernel potentials~\cite{mean} defined in terms of the feature vectors $f_p$ and $f_q$ for two voxels $v_p$ and $v_q$. The first term of this potential expresses the size and shape of the voxel-neighborhoods to encourage the homogeneous labels. The degree of this similarity is controlled by a parameter $\theta_\alpha$. This term is defined by
\begin{equation}
k^{(1)}(f_p,f_q)=w^{(1)}exp\left(-\sum_{d\in \{x,y\}}\frac{|v_{p,d}-v_{q,d}|^2}{2\theta^2_{\alpha,d}}\right)
\end{equation}
This kernel is defined by two diagonal covariance matrices (one for each axis) whose elements are the parameters $\theta_{\alpha,d}$.

The second term of the potential is similar, only an additional parameter $\gamma$ is used for interpreting how strong the homogeneous appearances of the voxels are in an area defined by spatial co-ordinates $d$ across the spectral channels $\lambda$. It is defined by
\begin{align}
k^{(2)}(f_p,f_q)=w^{(2)}exp (-\sum_{d\in \{x,y\}}\frac{|v_{p,d}-v_{q,d}|^2}{2\theta^2_{\alpha,d}} \nonumber \\-\sum_{\lambda \in B}\frac{|v_{p,\lambda}-v_{q,\lambda}|^2}{2\theta^2_{\gamma,\lambda}})
\end{align}
where $|v_{p,d}-v_{q,d}|$ is the spatial distance between voxels $p$ and $q$ and $|v_{p,\lambda}-v_{q,\lambda}|$ is their difference across the spectral domain. The influence of the unary and pairwise terms can be adjusted with their weights $w^{(1)}$ and $w^{(2)}$.

The inference algorithm in~\cite{mean} works in an iterative manner. The first step is initialization in which a soft-max function over the unary potential across all the labels for individual voxels is performed. The second step is message passing which applies convolutions with the two Gaussian kernels defined above on the current estimation of the prediction of the voxels. This reflects how strongly two voxels $v_p,v_q$ are related to each other. By using back propagation, we calculate error derivatives on the filter responses. The next step is to take the weighted sum of the filter outputs for each label of the voxels. When each label is considered, it can be reformulated as the usual convolution of filter with input voxels and the output labels. The error can be calculated since both inputs and outputs are known during the back-propagation. This allows an automatic learning of filter weights. Next, a compatibility transform step is performed followed by adding the original unary potential for each individual voxel obtained from the initial CNN. Finally, the normalization step of the iteration can be expressed as another softmax operation which gives us the final labels for the segments. Algorithm~\ref{deepcrfalgo} summarized the important stages of our deep CRF approach to segment the image.

\begin{algorithm}
\label{deepcrfalgo}
\caption{\color{blue}Deep CRF}
\color{blue}
\hspace*{\algorithmicindent} \textbf{Input:} 3D feature map obtained from Algorithm 1, M voxels in \{$v_1$,$v_2$,\dots,$v_M$\}

\begin{algorithmic}[1]
\FOR{Each $v$ in $M$}
\STATE Add $v$ in CRF graph
\FOR{Each $v_i$,$v_j$}
\IF{$v_i$ is adjacent to $v_j$}
\STATE Connect edge between $v_i$ and $v_j$ in CRF graph
\ENDIF
\ENDFOR
\ENDFOR
\FOR{Each $v$ in $M$}
        \STATE Compute unary potential function $\phi$ using Eq. (5)
        \ENDFOR
\FOR{Each $v_p$,$v_q$ in $N$}
        \STATE Compute pairwise potential function $\psi$ using Eq. (6)
        \ENDFOR
\STATE Compute two-kernel potentials using Eqs. (7) and (8)
\STATE Train CRF using Eqs. (12), (13) and (14)
\STATE Execute Mean-Field inference algorithm
\end{algorithmic}
    \hspace*{\algorithmicindent} \textbf{Output:} Segmented regions
\end{algorithm}

\textcolor[rgb]{0.00,0.00,1.00}{A segmentation map can be obtained from this step which suffers from low-resolution representation of inaccurate object boundaries due to repeated use of pooling layers during CNN training. To overcome this problem, we further employ deconvolution network during the CRF pairwise potential computation and produce refined output in the segmentation stage. We present the basic formulation of deconvolution and the reasons of using it in section~\ref{deconvoprediction}.}

\section{Prediction}
\label{deconvoprediction}

A convolution network has repeatedly used pooling layers in order to reduce the input image size which restricts the prediction performance of the network to some extent. It is not always possible to reconstruct a high-resolution representation of object boundaries accurately. As a result, low-resolution prediction of the original input image is produced in the prediction stage by CNN that eventually affects the CRF segmentation as well. To overcome this problem and further improve the prediction, we employed deconvolution network during the CRF pairwise potential computation to produce refined output in the segmentation stage.

\subsection{Prediction refinement with deconvolution network}
To obtain a high-resolution segmentation map from the mean-field inference, we add a deconvolution network~\cite{deconvo} into our framework. Although the use of deconvolution can be found in the literature, the learning of deconvolution is not very common. In our method, we learn a deep deconvolution network, which is composed of deconvolution, unpooling, and rectified linear unit (ReLU) layers~\cite{deconvo3}.

\subsubsection{Unpooling}
Pooling operation in a convolution network is very common. Pooling improves classification performance by filtering noisy activations in the lower layers and retaining activations in the upper layers only. It can abstract activations in a receptive field with a single value. Unfortunately, spatial information within a receptive field is lost during pooling. As a result, accurate localization that is required for image segmentation is not always possible. To overcome this problem, unpooling layers are used in the deconvolution network, which does the exact reverse operation of the pooling layers. During the CRF pairwise training, unpooling operation produces a finer resolution of an object by reconstructing the original size of the input data and thus restoring the detailed structures of the object of interest. Generally, unpooling operation keeps track of the locations of maximum activations which were selected during the pooling operation. This information can be very useful in placing the activations back to their original pooled location.

\subsubsection{Deconvolution}
The unpooling operation produces a large activation map which is not regular in nature. Although deconvolution operation is similar to convolution operations, it actually assigns a single input with multiple outputs unlike convolution operation which connects multiple inputs within a filter window or patch to a single activation value~\cite{deconvo,FCN}. This operation produces a much denser activation map compared to the sparse activation map obtained earlier. The filters used during deconvolution operation help in strengthening the activations that are close to the target classes and also suppressing the noisy activation from the regions containing different classes.  As a result, different layers of the deconvolution network can help in reconstructing shapes in different levels. The filters used in lower layers may help in reconstructing the overall shape of an object while the higher layer filters can help in more class-specific details of an object. Therefore, more refined and accurate segmentation outcome can be obtained by the use of deconvolution network.

In the proposed algorithm, we incorporate deconvolution into the CNN training only during the deep CRF training. This is because we want to produce a final dense segmentation map with class-specific information in it instead of simply applying it on low-resolution activation map as a separate step. The integration of deconvolution during the pairwise potential calculation of the deep CRF training particularly helps in improving the segmentation accuracy to a larger extent. The combination of unpooling, deconvolution, and rectification operations during the final CNN training of the deep CRF model contribute much in formulating the pairwise relations between image patches and improving the final segmentation performance.

\section{CRF Training}
Exact maximum-likelihood training for undirected graphical models is intractable as the computation involves marginal distribution calculation of the model. This is even more complex for conditional training when we are required to predict certain $l$ given observed input voxel $v$. This eventually leads the decision of optimizing $P(l|v)$ instead of $p(l,v)$. Parameter estimation in CRFs can be performed by maximizing the log likelihood of a training input-output pair ($v,l$) as previously defined in Equation~(\ref{CRF-model}) and Equation~(\ref{CRF-energy}). For the proposed CNN based CRF, the objective function for the CRF can be defined as
\begin{equation}
 \nabla(\theta)=\sum_{\substack{p\in M \\ (p,q) \in N \times N \\ \lambda \in B}} \phi(l_p,v_p)\psi(l_p,l_q,v_p,v_q)-Z(v;\theta_\lambda)
\end{equation}
Although such maximization of log-likelihood of ($v,l$) improves performance, the conditional training is expensive because the calculation of the log partition function $Z(v)$ depends on the model parameters as well as on the input voxels along the spectral channels. Therefore, estimating CRF parameters must include approximating the partition function for each iteration during the training phase in the stochastic gradient descent (SGD) method. This gets more complicated when a large number of iterative steps are required for SGD during the CNN training. Therefore, an efficient CRF training is desirable in order to reduce the computational complexities.

In order to efficiently train a large model, we can divide the entire model into pieces and then independently train those pieces. Later, we can combine the learned weights from those pieces and use it for testing purposes. This idea, known as piecewise training, was discussed in~\cite{piecewise}.

A proposition was defined and proved in~\cite{piecewise} about the piecewise estimator that maximizes a lower bound on the true likelihood. It says:
\begin{equation}
Z(A)\leq \sum_e Z(A|_e)
\label{proposition}
\end{equation}
Here, $A|_e$ is the vector A with zeros in the entries that do not correspond to the edge $e$. Therefore, the piecewise objective function for CRF can be defined for a training input-output pair ($v,l$) as:
\begin{equation}
\nabla(\theta)=\sum_{\substack{p\in M \\ (p,q) \in N \times N \\ \lambda \in B}} \phi(l_p,v_p)\psi(l_p,l_q,v_p,v_q)- \sum_{(p,q) \in N \times N} Z(v;\theta_\lambda)
\end{equation}
According to the proposition in Equation~(\ref{proposition}), for each $v$, the bound needs to be applied separately which removes the requirement of marginal inference for gradient calculation. This idea can be incorporated into CRF training with CNN potentials. We can formulate $P(l|v)$ as a number of independent likelihoods on both unary and pairwise potentials
\begin{equation}
P(l|v;\theta_\lambda)=\prod_\phi \prod_{p \in M} P_\phi(l_p|v;\theta) \prod_\psi \prod_{(p,q) \in N \times N} P_\psi(l_p,l_q|v;\theta_\lambda)
\end{equation}
Both $P_\phi(l_p|v;\theta_\lambda)$ and $P_\psi(l_p,l_q|v;\theta_\lambda)$ can be calculated from unary and pairwise potentials respectively.
\begin{equation}
P_\phi(l_p|v;\lambda)=\frac{exp[\phi(l_p,v;\theta_\lambda)]}{\sum_{l'_p}exp[\phi(l'_p,v;\theta_\lambda)]}
\end{equation}
\begin{equation}
P_\psi(l_p,l_q|v;\theta_\lambda)=\frac{exp[\psi(l_p,l_q,v;\theta_\lambda)]}{\sum_{l'_p,l'_q}exp[\psi(l'_p,l'_q,v;\theta_\lambda)]}
\end{equation}
As suggested by the piecewise CRF training evaluation function, it is not required to compute the partition function anymore and we only need to calculate the log likelihood of $P_\phi$ and $P_\psi$. As a result, the gradient calculation can be performed without partition function, thus saving expensive inference.

\section{Experimental Results}
In this section, we present the experimental results on real-world hyperspectral remote sensing images. Then we analyse the performance of the proposed method in comparison with several alternatives.

\subsection{Hyperspectral Image Datasets} In the experiments, we used two widely used hyperspectral datasets, i.e., Indian Pines and Pavia University, in order to evaluate the effectiveness of our proposed method.

\subsubsection{Indian Pines} Indian Pines data was acquired by the Airborne Visible/Infrared Spectrometer (AVIRIS). It consists of 145$\times$145 pixels and 220 spectral reflectance bands in the wavelength ranging from $0.4-2.5 \times 10^{-6}$ meters and has a spatial resolution of 20 meters. The number of bands was reduced to 200 by removing water absorption bands. Sixteen different classes on land-cover were considered in the ground truth.

\subsubsection{Pavia University} 
\begin{figure}[h]
\centering
\includegraphics[scale=1.25]{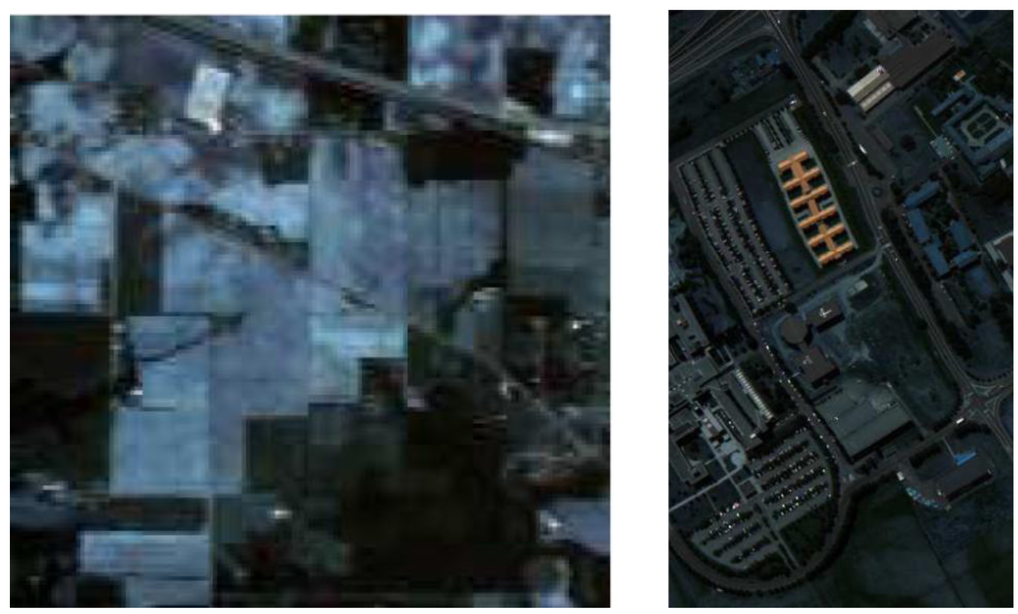}
\caption{Sample bands from (a) Indian Pines (b) Pavia University}
\label{Ind-Pav}
\end{figure}

Pavia University dataset was collected by Reflective Optics System Imaging Spectrometer (ROSIS-3). It consists of 610$\times$340 pixels and 115 spectral reflectance bands in the wavelength ranging from $0.43-0.86 \times 10^{-6}$ meters and has a spatial resolution of 1.3 meters. The number of bands was reduced to 103 by removing noisy bands. Nine different classes on land-cover were considered in the ground truth. Fig.~\ref{Ind-Pav} shows randomly chosen bands extracted from Indian Pines and Pavia University datasets.

\begin{figure}[h]
\centering
\begin{tabular}{cc}
    \includegraphics[width=0.45\linewidth]{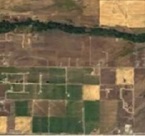}&
    \includegraphics[width=0.45\linewidth]{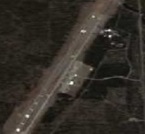}\\[2\tabcolsep] \\
    \end{tabular}
        \caption{Image instances from the new dataset Griffith-USGS}
        \label{newDS}
\end{figure}

\subsubsection{A New Dataset} For better evaluation of our proposed method, we created a new dataset by collecting AVIRIS images from the USGS database\footnote{https://earthexplorer.usgs.gov/}. After downloading data available in the USGS website, we performed a subsequent pre-processing step to make the images compatible for use in hyperspectral image analysis. After that, we manually labeled the images. The details on the construction of this dataset are described in Section~\ref{sec:newdataset}.

We separated our experiments into two stages: classification and segmentation. For both tasks, we compared our results with state-of-the-art methods to evaluate the usefulness of our proposed algorithm. The details of our experiments will be presented later.

\subsection{Construction of the new dataset}\label{sec:newdataset}
In the official AVIRIS website\footnote{https://aviris.jpl.nasa.gov/alt\_locator/}, we downloaded remote sensing data located in the region of north America spanning over the United States of America, Canada and Mexico. We used the data acquisition tool provided by the website for selecting the regions from where data need to be extracted. The AVIRIS sensor collects data that can be used for characterization of the Earth's surface and atmosphere from geometrically coherent spectroradiometric measurements. With proper calibration and correction for atmospheric effects, the measurements can be converted to ground reflectance data which then can be used for quantitative characterization of surface features.

In our research, we downloaded 19 scenes to build the training and testing sets for deep learning. We cropped those scenes into a number of individual portions to build 25 training images and 35 testing images. As we captured scenes from multiple locations, the spatial resolutions of the scenes used in this dataset range from 2.4 to 18 meters. Each image consists of approximately 145$\times$145 pixels. Fig.~\ref{newDS} shows two instances from the training set of our new dataset \emph{Griffith-USGS}.

\subsection{Pre-processing}
After collecting the AVIRIS image data, the following step was to undertake some pre-processing tasks in order to convert images into a suitable form for proper use. Hyperspectral sensors should be spectrally and radiometrically calibrated before data analysis. NASA/JPL has already processed the AVIRIS data to remove geometric and radiometric errors associated with the motion of the aircraft during data collection. However, the data should be further corrected for atmospheric effects and converted to surface reflectance prior to scientific analysis. To do the conversion, we used a tool `'FLAASH'' \cite{FLAASH} provided by ENVI. FLAASH is a model-based radiative transfer program to convert radiance data to reflectance. Developed by Spectral Sciences, Inc., FLAASH uses MODTRAN4 radiation transfer code to correct images.

\subsection{Manual Labelling}
After obtaining the reflectance data, the next step was to create the training and testing datasets accordingly. The data that we obtained from AVIRIS were not labeled. As our method relies on a supervised training approach, it was important to construct a labeled set in order to fit into our proposed framework. For this purpose, we performed a pixelwise manual labeling on the images. To increase the size of the training set, we cropped smaller portions from the original image. We made sure that the cropped portion should contain instances of at least few classes that we want to segment. We created a training set containing six classes, including road, water, building, grass, tree and soil. The labeling was done with the help of high-resolution color images in Google Earth.

\subsection{Design of the CNNs}
As mentioned before, we used spectral cube-based CNN in this method that learns both spectral and spatial features. In this section, we elaborate the design of all the CNNs used in various stages of our framework.

For each CNNs used in our method, we used 32 $5 \times 5 \times 5$  convolution kernels. Depending on the datasets and the classification and segmentation stages, we adopted four to seven convolution layers and two to four pooling layers with $2 \times 2$ pooling kernel in each layer. The analysis on the selection of convolution layers is provided later. ReLU layers were used for all datasets as well. All layers were trained using backpropagation/SGD. The architecture of the CNNs used in our method is explained in Table \ref{architecture}.

\begin{table}[]
\centering
\caption{Architecture of the CNNs}
\label{architecture}
\begin{tabular}{|l|l|l|l|l}
\hline
\multirow{2}{*}{Dataset}                                & \multicolumn{2}{l|}{Classification}                                                                                                 & \multicolumn{2}{l|}{Segmentation}                                                                                                                                                            \\ \cline{2-5}
                                                        & \multicolumn{1}{l|}{Layer}                                                                                                & Pooling & Layer                                                                           & \multicolumn{1}{l|}{Pooling}                                                                               \\ \hline
\multirow{7}{*}{Indian Pines}                           & \multicolumn{1}{l|}{\multirow{7}{*}{\begin{tabular}[c]{@{}l@{}}1\\ 2\\  3\\  4\\  5\\  6\\  7\end{tabular}}} & 2$\times$2     & \multirow{4}{*}{\begin{tabular}[c]{@{}l@{}}1\\  2\\  3\\ 4\end{tabular}} & \multicolumn{1}{l|}{\multirow{4}{*}{\begin{tabular}[c]{@{}l@{}}2$\times$2\\  No\\  No\\  2$\times$2\end{tabular}}} \\
                                                        & \multicolumn{1}{l|}{}                                                                                                     & 2$\times$2     &                                                                                 & \multicolumn{1}{l|}{}                                                                                      \\
                                                        & \multicolumn{1}{l|}{}                                                                                                     & No      &                                                                                 & \multicolumn{1}{l|}{}                                                                                      \\
                                                        & \multicolumn{1}{l|}{}                                                                                                     & No      &                                                                                 & \multicolumn{1}{l|}{}                                                                                      \\
                                                        & \multicolumn{1}{l|}{}                                                                                                     & 2$\times$2     &                                                                                 & \multicolumn{1}{l|}{}                                                                                      \\
                                                        & \multicolumn{1}{l|}{}                                                                                                     & No      &                                                                                 & \multicolumn{1}{l|}{}                                                                                      \\
                                                        & \multicolumn{1}{l|}{}                                                                                                     & 2$\times$2     &                                                                                 & \multicolumn{1}{l|}{}                                                                                      \\ \cline{1-2} \cline{2-5}
\multirow{7}{*}{Pavia University}                           & \multicolumn{1}{l|}{\multirow{6}{*}{\begin{tabular}[c]{@{}l@{}}1\\ 2\\  3\\  4\\  5\\  6\end{tabular}}} & 2$\times$2     & \multirow{3}{*}{\begin{tabular}[c]{@{}l@{}}1\\  2\\  3\end{tabular}} & \multicolumn{1}{l|}{\multirow{3}{*}{\begin{tabular}[c]{@{}l@{}}2$\times$2\\  No\\  2$\times$2\end{tabular}}} \\
                                                        & \multicolumn{1}{l|}{}                                                                                                     & 2$\times$2     &                                                                                 & \multicolumn{1}{l|}{}                                                                                      \\
                                                        & \multicolumn{1}{l|}{}                                                                                                     & No      &                                                                                 & \multicolumn{1}{l|}{}                                                                                      \\
                                                        & \multicolumn{1}{l|}{}                                                                                                     & No      &                                                                                 & \multicolumn{1}{l|}{}                                                                                      \\
                                                        & \multicolumn{1}{l|}{}                                                                                                     & 2$\times$2     &                                                                                 & \multicolumn{1}{l|}{}                                                                                      \\
                                                        & \multicolumn{1}{l|}{}                                                                                                     & No      &                                                                                 & \multicolumn{1}{l|}{}                                                                                      \\
                                                        & \multicolumn{1}{l|}{}                                                                                                     &      &                                                                                 & \multicolumn{1}{l|}{}                                                                                      \\ \cline{1-2} \cline{2-5}

\multirow{6}{*}{Griffith-USGS}                           & \multicolumn{1}{l|}{\multirow{6}{*}{\begin{tabular}[c]{@{}l@{}}1\\ 2\\  3\\  4\\  5\\  6\end{tabular}}} & 2$\times$2     & \multirow{3}{*}{\begin{tabular}[c]{@{}l@{}}1\\  2\\  3\end{tabular}} & \multicolumn{1}{l|}{\multirow{3}{*}{\begin{tabular}[c]{@{}l@{}}2$\times$2\\  No\\  No\end{tabular}}} \\
                                                        & \multicolumn{1}{l|}{}                                                                                                     & 2$\times$2     &                                                                                 & \multicolumn{1}{l|}{}                                                                                      \\
                                                        & \multicolumn{1}{l|}{}                                                                                                     & No      &                                                                                 & \multicolumn{1}{l|}{}                                                                                      \\
                                                        & \multicolumn{1}{l|}{}                                                                                                     & No      &                                                                                 & \multicolumn{1}{l|}{}                                                                                      \\
                                                        & \multicolumn{1}{l|}{}                                                                                                     & 2$\times$2     &                                                                                 & \multicolumn{1}{l|}{}                                                                                      \\
                                                        & \multicolumn{1}{l|}{}                                                                                                     & No      &                                                                                 & \multicolumn{1}{l|}{}                                                                                      \\
                                                        & \multicolumn{1}{l|}{}                                                                                                     &      &                                                                                 & \multicolumn{1}{l|}{}                                                                                      \\ \cline{1-2} \cline{2-5}
\end{tabular}
\end{table}

The ReLU layer cuts off the features that are less than 0. The pooling layers reduces the size of the feature maps. The size of the mini-batch is set to 100. For the logistic regression, the learning rate is set to 0.003 for Indian Pines, 0.01 for Pavia University and 0.005 for our new dataset. The number of epochs was 700 in classification and 500 in segmentation for Indian Pines, 600 in classification and 500 in segmentation for Pavia University and for Griffith-USGS. The weights are randomly initialized and gradually trained using the back propagation algorithm. Each convolution kernel extracts distinct features from the input and with the weights being learned. The features convey meaningful structural information about the data. Different kernels used in the convolution layers are able to extract different features on the way to form a powerful representation.

\subsection{Results and Comparisons}
\textcolor[rgb]{0.00,0.00,1.00}{In this paper, we intended to establish a meaningful connection between the tasks of classification and segmentation. The increasing demands of critical image analysis tasks such as improving the quality of the segmentation results introduce the importance of employing multiple cues regarding the image structure. In this regard, pixel-wise classification provides useful input for the task of segmentation in terms of generating an effective spectral-spatial structure representation of the hyperspectral data. These initial inputs guide the subsequent process of constructing deep CRF for segmentation. It is, in fact, a common practice to use classification as an initial step for performing segmentation~\cite{Li, seg-class}. It is also a common practice to compare the accuracy of classification and segmentation in hyperspectral remote sensing~\cite{spatial, Li}.}

To begin with, we evaluated the effectiveness of the first part of our framework which includes the execution of CNN over the spectral cubes of the image that eventually results in the classification step. For this purpose, we compared this part to an SVM-based classification algorithm~\cite{composite} which itself is divided into two parts: (1) SVM with composite kernel (SVM-CK) and (2) SVM with generalized composite kernel (SVM-GCK). We compared our results with SVM-GCK as it outperformed its counterpart SVM-CK.

We also compared our method with a spatial-spectral-based method (MPM-LBP-AL)~\cite{spatial}. In this method, active learning (AL) and loopy belief propagation (LBP) algorithms were used to learn spectral and spatial information simultaneously. Then the marginal probability distribution were exploited, which used the whole information in the hyperspectral data. We made comparisons with another supervised method (MLR\textit{sub}MLL)~\cite{Li} that integrated spectral and spatial information into a Bayesian framework. In this method, a multinomial logistic regression (MLR) algorithm was used to learn the posterior probability distributions from the spectral information. Moreover, a subspace projection method was used to characterize noisy and mixed pixels. Later, spatial information is added using a multilevel logistic MRF prior. Along with this, we also reported the performance of a recent work developed by Chen \emph{et al.}~\cite{yushi} who proposed classification methods based on 1-D, 2-D and 3-D CNNs. To fit into our method, we simply compared with their 1-D CNN (1D-CNN-LR) and 3-D CNN (3D-CNN-LR) approaches which used logistic regression (LR) to classify pixels.

\textcolor[rgb]{0.00,0.00,1.00}{We chose limited samples for training since we want to simulate the real-world cases where the size of labelled data is small. For our experiments, we chose 3 samples per class in the extreme case and continued investigating on different numbers of training samples per class, from 5 to 15. To improve the classification performance and to avoid overfitting problem, we increased the size of the training samples by augmentation discussed in section~\ref{virtual-sample}. We used those limited real samples for augmenting the training set and the rest of the real samples were included in the testing set. Classification performance varied to a significant extent according to the selected training samples since samples with good/bad representations affected the performance. Also, during the CNN training, we used 90\% of the total number of training samples, consisting of limited number of real samples and augmented samples, to learn the weights and biases of the neurons and the rest of the 10\% to validate and further update the design of the architecture.}

\textcolor[rgb]{0.00,0.00,1.00}{For performance evaluation, we calculated the overall accuracy (OA) and average accuracy (AA) with the corresponding standard deviations. We repeated our experiments for ten times over the randomly split training and testing data. Furthermore, we assessed the statistical significance of our results by applying binomial test in which the assessment was done by computing the \textit{p}-value from the paired \textit{t}-test. We set the confidence interval to 95\% which declares statistical significance at $p<.05$ level.}
\begin{table*}[t]
\centering
\caption{Classification Accuracies on Different Datasets (Pixelwise). A `*' denotes that the best average accuracy (shown in bold) is significantly better that the accuracy achieved by the corresponding method according to a statistical paired t-test for comparing classifiers}
\label{Classification-Pixelwise}
\begin{tabular}{|c|c|c|c|c|c|c|}
\hline
\begin{tabular}[c]{@{}c@{}}Dataset\end{tabular} & \begin{tabular}[c]{@{}l@{}}\end{tabular} &  \begin{tabular}[c]{@{}l@{}}SVM-GCK~\cite{composite}\end{tabular} & \begin{tabular}[c]{@{}c@{}}MPM-LBP-AL~\cite{spatial}\end{tabular} & \begin{tabular}[c]{@{}l@{}}MLR\textit{sub}MLL~\cite{Li}\end{tabular} & \begin{tabular}[c]{@{}c@{}}1D-CNN-LR~\cite{yushi}\end{tabular} & \begin{tabular}[c]{@{}l@{}}Proposed Method\end{tabular} \\  \hline
\multirow{2}{*}{Indian Pines}     & OA (\%) & 87.53 $\pm$ 2.30     & 90.07 $\pm$ 1.76        & 85.06 $\pm$ 1.92       & \textbf{92.93 $\pm$ 1.44} & 92.59 $\pm$ 0.55          \\ \cline{2-7}
                                  & AA (\%) & 88.97 $\pm$ 0.54$^*$     & 90.01 $\pm$ 0.77        & 86.00 $\pm$ 1.09$^*$       & \textbf{93.05 $\pm$ 2.14} & 92.96 $\pm$ 1.01          \\ \hline
\multirow{2}{*}{Pavia University} & OA (\%) & 89.39 $\pm$ 2.19     & 84.70 $\pm$ 1.22        & 87.97 $\pm$ 1.54       & \textbf{92.35 $\pm$ 1.08} & 92.06 $\pm$1.36           \\ \cline{2-7}
                                  & AA (\%) & 91.98 $\pm$ 1.23     & 85.97 $\pm$ 0.07$^*$        & 89.31 $\pm$ 0.77$^*$       & 93.17 $\pm$ 1.26          & \textbf{93.97 $\pm$ 0.30} \\ \hline
\multirow{2}{*}{Griffith-USGS}    & OA (\%) & 67.33 $\pm$ 2.71     & 68.69 $\pm$ 0.91        & 68.05 $\pm$ 0.19       & 75.07 $\pm$ 1.23          & \textbf{75.97 $\pm$ 0.19} \\ \cline{2-7}
                                  & AA (\%) & 70.45 $\pm$ 1.49$^*$     & 69.33 $\pm$ 1.01$^*$        & 69.02 $\pm$ 0.77$^*$       & 75.98$\pm$ 1.30$^*$           & \textbf{76.42 $\pm$ 0.83} \\ \hline
\end{tabular}
\end{table*}
\textcolor[rgb]{0.00,0.00,1.00}{Table \ref{Classification-Pixelwise} reports the pixelwise CNN-based classification results on Indian Pines, Pavia University and Griffith-USGS datasets with 15 real samples per class and augmented samples for training. The results show that our method achieved similar accuracy as 1D-CNN-LR~\cite{yushi}. Both methods outperformed other pixel-wise classification methods and were statistically significant in most cases. Therefore, we can conclude that the CNN-based approaches can effectively improve the classification accuracy.}

\begin{table*}[t]
\centering
\caption{Classification Accuracies on Different Datasets (Spectral Cubes). A `*' denotes that the best average accuracy (shown in bold) is significantly better than the accuracy achieved by the corresponding method according to a statistical paired t-test for comparing classifiers.}
\label{Classification-Cubes}
\begin{tabular}{|c|c|c|c|c|c|c|}
\hline
\begin{tabular}[c]{@{}c@{}}Dataset\end{tabular} & \begin{tabular}[c]{@{}l@{}}\end{tabular} &  \begin{tabular}[c]{@{}l@{}}SVM-GCK~\cite{composite}\end{tabular} & \begin{tabular}[c]{@{}c@{}}MPM-LBP-AL~\cite{spatial}\end{tabular} & \begin{tabular}[c]{@{}l@{}}MLR\textit{sub}MLL~\cite{Li}\end{tabular} & \begin{tabular}[c]{@{}c@{}}3D-CNN-LR~\cite{yushi}\end{tabular} & \begin{tabular}[c]{@{}l@{}}Proposed Method\end{tabular} \\  \hline
\multirow{2}{*}{Indian Pines}     & OA (\%) & 90.70 $\pm$ 1.35     & 92.20 $\pm$ 1.82        & 90.66 $\pm$ 0.20       & 97.88 $\pm$ 0.48       & \textbf{98.29 $\pm$ 0.33} \\ \cline{2-7}
                                  & AA (\%) & 90.83 $\pm$ 0.32$^*$     & 92.18 $\pm$ 1.21$^*$        & 89.91 $\pm$ 2.30$^*$       & 99.18 $\pm$ 0.06       & \textbf{99.20 $\pm$ 0.09} \\ \hline
\multirow{2}{*}{Pavia University} & OA (\%) & 96.14 $\pm$ 2.19     & 87.25 $\pm$ 1.26        & 93.91 $\pm$ 1.44       & 98.60 $\pm$ 0.07       & \textbf{99.12 $\pm$ 0.41} \\ \cline{2-7}
                                  & AA (\%) & 96.05 $\pm$ 0.11     & 89.09 $\pm$ 0.08$^*$        & 92.00 $\pm$ 1.04$^*$       & 99.53 $\pm$ 0.05       & \textbf{99.69 $\pm$ 0.03} \\ \hline
\multirow{2}{*}{Griffith-USGS}    & OA (\%) & 73.97 $\pm$ 1.21     & 63.19 $\pm$ 1.99        & 68.88 $\pm$ 1.45       & 77.71 $\pm$ 0.87       & \textbf{83.05 $\pm$ 1.19} \\ \cline{2-7}
                                  & AA (\%) & 74.97 $\pm$ 0.46$^*$     & 65.02 $\pm$ 0.97$^*$        & 69.95 $\pm$ 1.45$^*$       & 78.95 $\pm$ 0.37$^*$       & \textbf{84.98 $\pm$ 0.86} \\ \hline
\end{tabular}
\end{table*}
\textcolor[rgb]{0.00,0.00,1.00}{As described before, we proposed to perform CNN-based classification on spectral cubes instead of pixel-wise classification. Table~\ref{Classification-Cubes} reports 3-D CNN-based classification results on Indian Pines, Pavia University and Griffith-USGS datasets with 15 samples per class and augmented samples for training. By keeping the CNN parameters the same as in the pixel-based classification experiments, it is evident from the results in Tables ~\ref{Classification-Pixelwise} and~\ref{Classification-Cubes} that the classification accuracy can be significantly improved with spectral cube-based representation. The reason is that 3D operation better characterizes the spatial and structural properties of the hyperspectral data. Both our method and 1D-CNN-LR~\cite{yushi} outperform the rest of the methods, showing the power of deep neural networks. Our method significantly outperforms 1D-CNN-LR~\cite{yushi} on Griffith-USGS dataset, which proves the usefulness of the proposed paradigm.}

\textcolor[rgb]{0.00,0.00,1.00}{Since it is a common practice to compare the accuracy of classification and segmentation in hyperspectral remote sensing~\cite{spatial, Li}, we evaluated the effectiveness of the later stages of our framework with MLR\textit{sub}MLL~\cite{Li} and WHED~\cite{Watershed} which also included explicit segmentation stages. We further report the improved accuracy of 3D-CNN-LR~\cite{yushi} since this method included L2 regularization and dropout in the training process to improve the initial coarse classification results and produce more refined output. Similarly, we also report the final accuracies obtained by MPM-LBP-AL~\cite{spatial} in which active learning was used in the later stages of their algorithm to improve the accuracy previously obtained by estimating the marginal inference for the whole image.  We tested the usefulness of deconvolution by (1) running the method without using deconvolution at all, (2) using deconvolution during CRF unary potential calculation and (3) using deconvolution during CRF pairwise potential calculation stage. It is important to note that we prefer to include deconvolution into pairwise potential calculation stage as this step plays a major role in constructing accurate segments by connecting regions that actually belong to the same segment. Therefore, we applied deconvolution in the deep pairwise potential calculation stage rather than using it in other stages.}

\begin{table*}[t]
\centering
\caption{\color{blue}Segmentation Accuracies on Different datasets. A `*' denotes that the best average accuracy (shown in bold) is significantly better than the accuracy achieved by the corresponding method according to a statistical paired t-test for comparing classifiers.}
\label{SegmentationFinal}
\begin{tabular}{|c|c|c|c|c|c|c|c|c|}
\hline
{\color[HTML]{3531FF} }                                   & {\color[HTML]{3531FF} }                   & {\color[HTML]{3531FF} }                                                                          & {\color[HTML]{3531FF} }                                                                         & {\color[HTML]{3531FF} }                       & {\color[HTML]{3531FF} }                                                                         & \multicolumn{3}{c|}{{\color[HTML]{3531FF} Proposed Method}}                                                                                                                                                                                                                           \\ \cline{7-9} 
\multirow{-2}{*}{{\color[HTML]{3531FF} Dataset}}          & \multirow{-2}{*}{{\color[HTML]{3531FF} }} & \multirow{-2}{*}{{\color[HTML]{3531FF} \begin{tabular}[c]{@{}c@{}}MPM-LBP-AL\\~\cite{spatial}\end{tabular}}} & \multirow{-2}{*}{{\color[HTML]{3531FF} \begin{tabular}[c]{@{}c@{}}MLRsubMLL\\~\cite{Li}\end{tabular}}} & \multirow{-2}{*}{{\color[HTML]{3531FF} WHED~\cite{Watershed}}} & \multirow{-2}{*}{{\color[HTML]{3531FF} \begin{tabular}[c]{@{}c@{}}3D-CNN-LR\\~\cite{yushi}\end{tabular}}} & {\color[HTML]{3531FF} \begin{tabular}[c]{@{}c@{}}Without\\ Deconvolution\end{tabular}} & {\color[HTML]{3531FF} \begin{tabular}[c]{@{}c@{}}Deconvolution\\ in Unary CRF\end{tabular}} & {\color[HTML]{3531FF} \begin{tabular}[c]{@{}c@{}}Deconvolution\\ in Pairwise CRF\end{tabular}} \\ \hline
{\color[HTML]{3531FF} }                                   & {\color[HTML]{3531FF} OA (\%)}            & {\color[HTML]{3531FF} 92.91 $\pm$ 1.24}                                                          & {\color[HTML]{3531FF} 91.85 $\pm$ 0.83}                                                         & {\color[HTML]{3531FF} 90.15 $\pm$ 1.95}       & {\color[HTML]{3531FF} 98.25 $\pm$ 0.78}                                                         & {\color[HTML]{3531FF} 98.38 $\pm$ 0.37}                                                & {\color[HTML]{3531FF} 99.04 $\pm$ 0.03}                                                     & {\color[HTML]{3531FF} \textbf{99.15 $\pm$ 0.16}}                                             \\ \cline{2-9} 
\multirow{-2}{*}{{\color[HTML]{3531FF} Indian Pines}}     & {\color[HTML]{3531FF} AA (\%)}            & {\color[HTML]{3531FF} 92.35 $\pm$ 1.90$^*$}                                                      & {\color[HTML]{3531FF} 91.95 $\pm$ 0.74$^*$}                                                     & {\color[HTML]{3531FF} 90.85 $\pm$ 2.05$^*$}   & {\color[HTML]{3531FF} 99.27 $\pm$ 0.12}                                                         & {\color[HTML]{3531FF} 99.29 $\pm$ 0.24}                                                & {\color[HTML]{3531FF} 99.35 $\pm$ 0.10}                                                     & {\color[HTML]{3531FF} \textbf{99.41 $\pm$ 0.04}}                                             \\ \hline
{\color[HTML]{3531FF} }                                   & {\color[HTML]{3531FF} OA (\%)}            & {\color[HTML]{3531FF} 92.19 $\pm$ 0.50}                                                          & {\color[HTML]{3531FF} 94.77 $\pm$ 1.09}                                                         & {\color[HTML]{3531FF} 87.85 $\pm$ 1.75}       & {\color[HTML]{3531FF} 98.80 $\pm$ 0.28}                                                         & {\color[HTML]{3531FF} 99.23 $\pm$ 0.03}                                                & {\color[HTML]{3531FF} 99.32 $\pm$ 0.13}                                                     & {\color[HTML]{3531FF} \textbf{99.63 $\pm$ 0.07}}                                             \\ \cline{2-9} 
\multirow{-2}{*}{{\color[HTML]{3531FF} Pavia University}} & {\color[HTML]{3531FF} AA (\%)}            & {\color[HTML]{3531FF} 93.85 $\pm$ 0.16$^*$}                                                      & {\color[HTML]{3531FF} 95.35 $\pm$ 0.71$^*$}                                                     & {\color[HTML]{3531FF} 86.50 $\pm$ 2.56$^*$}   & {\color[HTML]{3531FF} 99.60 $\pm$ 0.07}                                                         & {\color[HTML]{3531FF} 99.63 $\pm$ 0.05}                                                & {\color[HTML]{3531FF} 99.70 $\pm$ 0.06}                                                     & {\color[HTML]{3531FF} \textbf{99.79 $\pm$ 0.03}}                                             \\ \hline
{\color[HTML]{3531FF} }                                   & {\color[HTML]{3531FF} OA (\%)}            & {\color[HTML]{3531FF} 69.89 $\pm$ 1.47}                                                          & {\color[HTML]{3531FF} 74.29 $\pm$ 0.66}                                                         & {\color[HTML]{3531FF} 70.20 $\pm$ 2.33}       & {\color[HTML]{3531FF} 82.53 $\pm$ 0.69}                                                         & {\color[HTML]{3531FF} 84.91 $\pm$ 1.02}                                                & {\color[HTML]{3531FF} 86.00 $\pm$ 0.29}                                                     & {\color[HTML]{3531FF} \textbf{88.92 $\pm$ 0.17}}                                             \\ \cline{2-9} 
\multirow{-2}{*}{{\color[HTML]{3531FF} Griffith-USGS}}    & {\color[HTML]{3531FF} AA (\%)}            & {\color[HTML]{3531FF} 70.19 $\pm$ 2.47$^*$}                                                      & {\color[HTML]{3531FF} 75.06 $\pm$ 1.20$^*$}                                                     & {\color[HTML]{3531FF} 71.85 $\pm$ 2.50$^*$}   & {\color[HTML]{3531FF} 83.04 $\pm$ 0.91$^*$}                                                     & {\color[HTML]{3531FF} 84.55 $\pm$ 1.36$^*$}                                            & {\color[HTML]{3531FF} 85.05 $\pm$ 1.05$^*$}                                                 & {\color[HTML]{3531FF} \textbf{89.13 $\pm$ 0.74}}                                             \\ \hline
\end{tabular}
\end{table*}
Table \ref{SegmentationFinal} reports the segmentation accuracies on three datasets respectively. The results show that our algorithm outperforms the methods MLR\textit{sub}MLL~\cite{Li}, MPM-LBP-AL~\cite{spatial}, 3D-CNN-LR~\cite{yushi} and WHED~\cite{Watershed}. \textcolor[rgb]{0.00,0.00,1.00}{The integrated CNN-based pairwise potentials defined on both spatial and spectral dimensions significantly improved the coarse-level prediction rather than doing local smoothness. During our experiments, we observed that the classification map produced by the initial CNN was too coarse for the Griffith-USGS dataset since we collected images from different scenes. Those scenes varied significantly in terms of resolution and contrast, and hence introduced more challenges in producing a refined segmentation map. After integrating CRF potentials, an approximately 7\% increase in accuracy was observed (Table IV), leading to significant advantages over the baseline methods.}  Deconvolution network is capable of improving of the final output, particularly when it is used during the pairwise potential calculation, by effectively improving the accuracy on connecting regions that belong to the same segment. The idea of integrating deconvolution into pairwise potential computation was supported by the results this option outperforms the other two versions where deconvolution was not used at all and was used in calculating unary potentials. The results are presented in Table~\ref{deconvo}.
\begin{table}[t]
\centering
\caption{Effect of Deconvolution.}
\label{deconvo}
\begin{tabular}{|c|c|c|c|}
\hline
\multirow{3}{*}{Dataset} & \multicolumn{3}{c|}{Accuracy (\%)}                                                                                                                                                                                     \\ \cline{2-4}
                         & \begin{tabular}[c]{@{}c@{}}Without\\ Deconvolution\end{tabular} & \begin{tabular}[c]{@{}c@{}}Deconvolution \\ in Unary CRF\end{tabular} & \begin{tabular}[c]{@{}c@{}}Deconvolution\\ in Pairwise CRF\end{tabular} \\ \hline
Indian Pines             & 99.29 & 99.35                                                                 & \textbf{99.41}                                                                   \\ \hline
Pavia University         & 99.63                                                           & 99.70                                                                 & \textbf{99.79}                                                                   \\ \hline
Griffith-USGS              & 85.55                                                           & 86.92                                                                 & \textbf{89.13}                                                                   \\ \hline
\end{tabular}
\end{table}

\begin{figure*}[t]
\centering
\begin{tabular}{ccccc}

\includegraphics[width=0.13\linewidth]{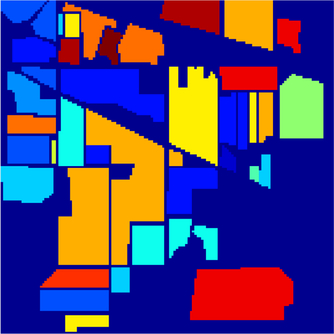}&
      \includegraphics[width=0.13\linewidth]{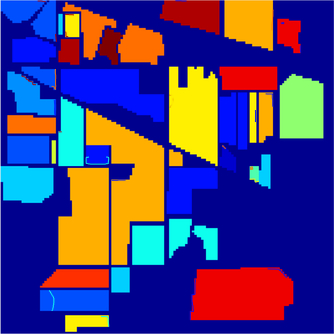}&
       \includegraphics[width=0.13\linewidth]{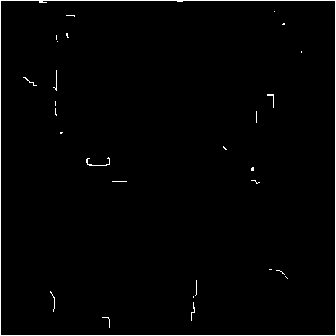}&
        \includegraphics[width=0.13\linewidth]{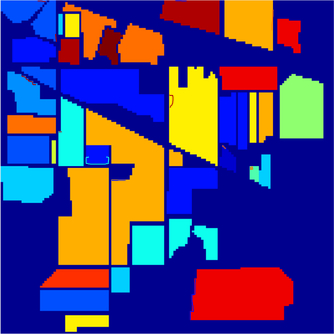}&
    \includegraphics[width=0.13\linewidth]{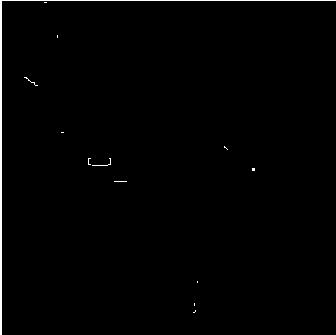}\\[2\tabcolsep]

    \includegraphics[width=0.13\linewidth]{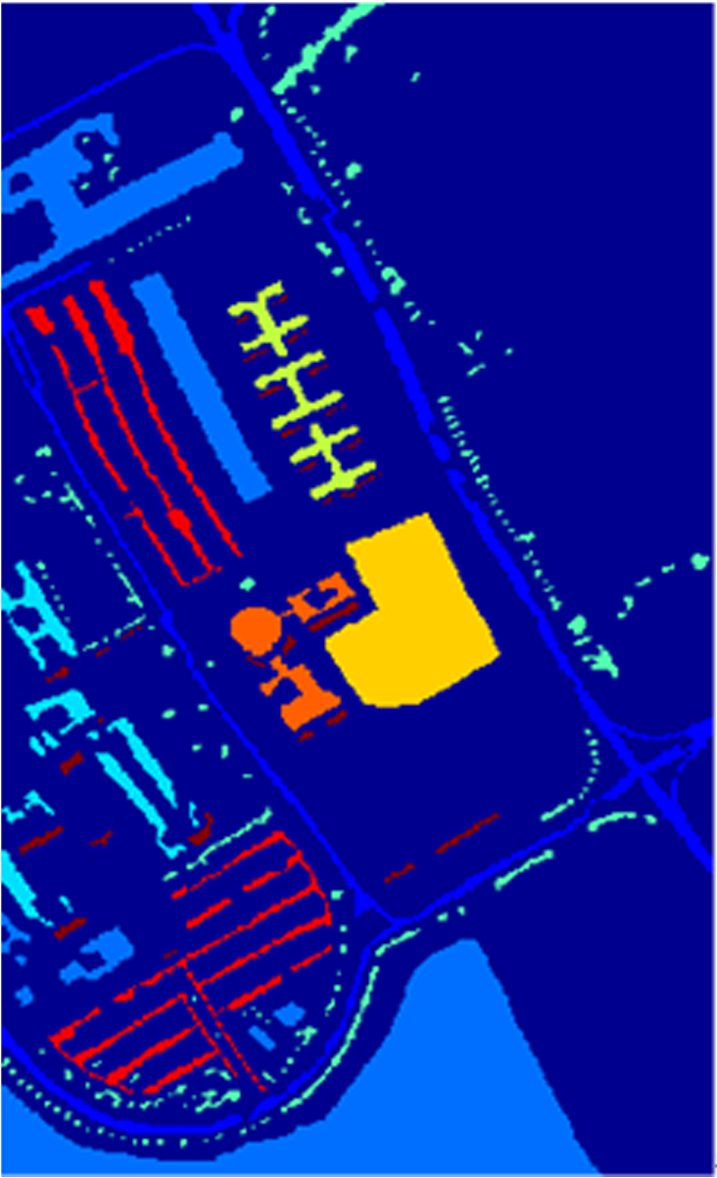}&
      \includegraphics[width=0.13\linewidth]{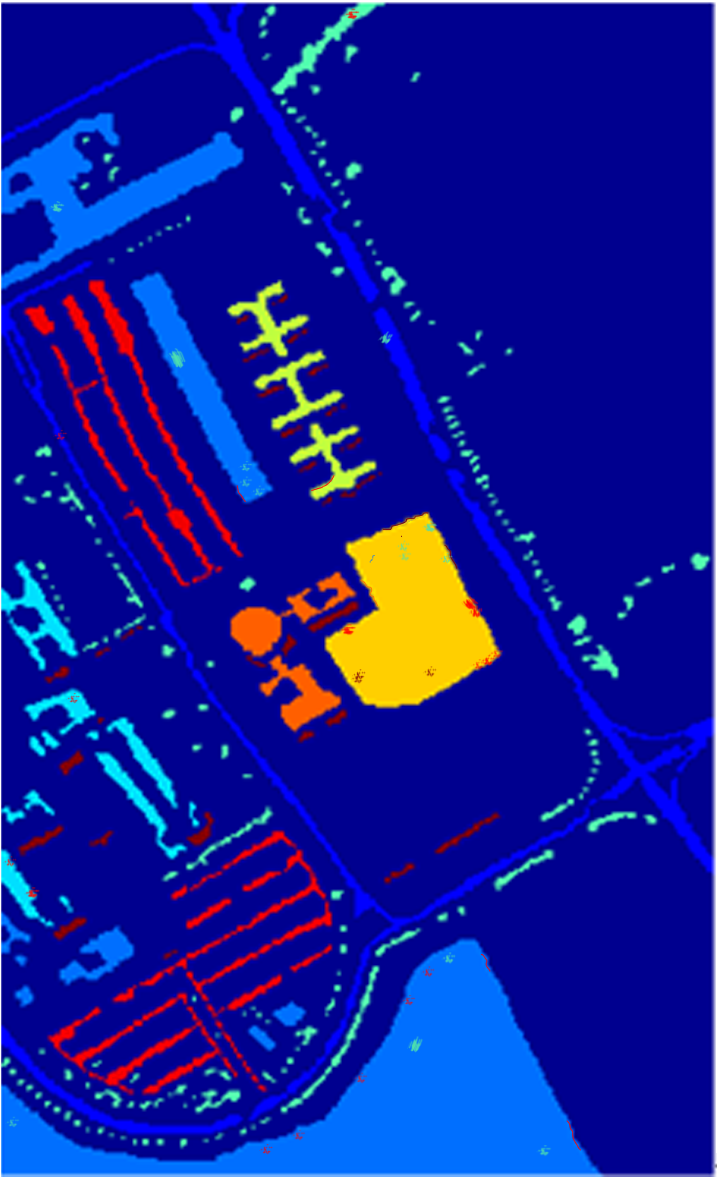}&
       \includegraphics[width=0.13\linewidth]{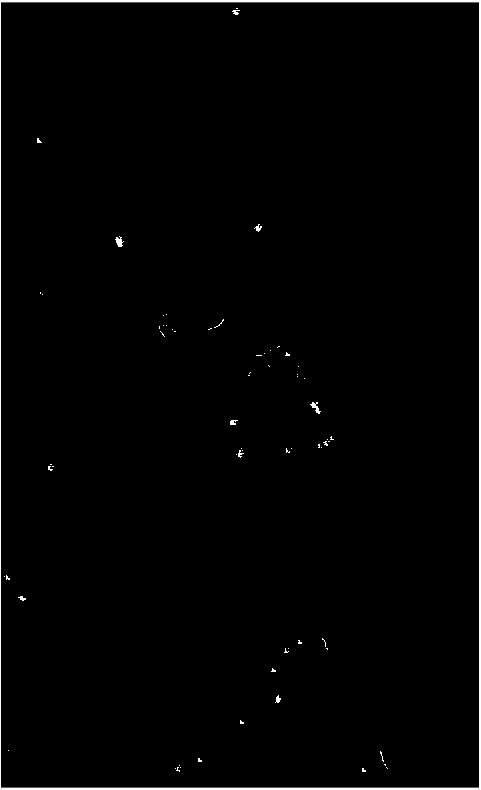}&
        \includegraphics[width=0.13\linewidth]{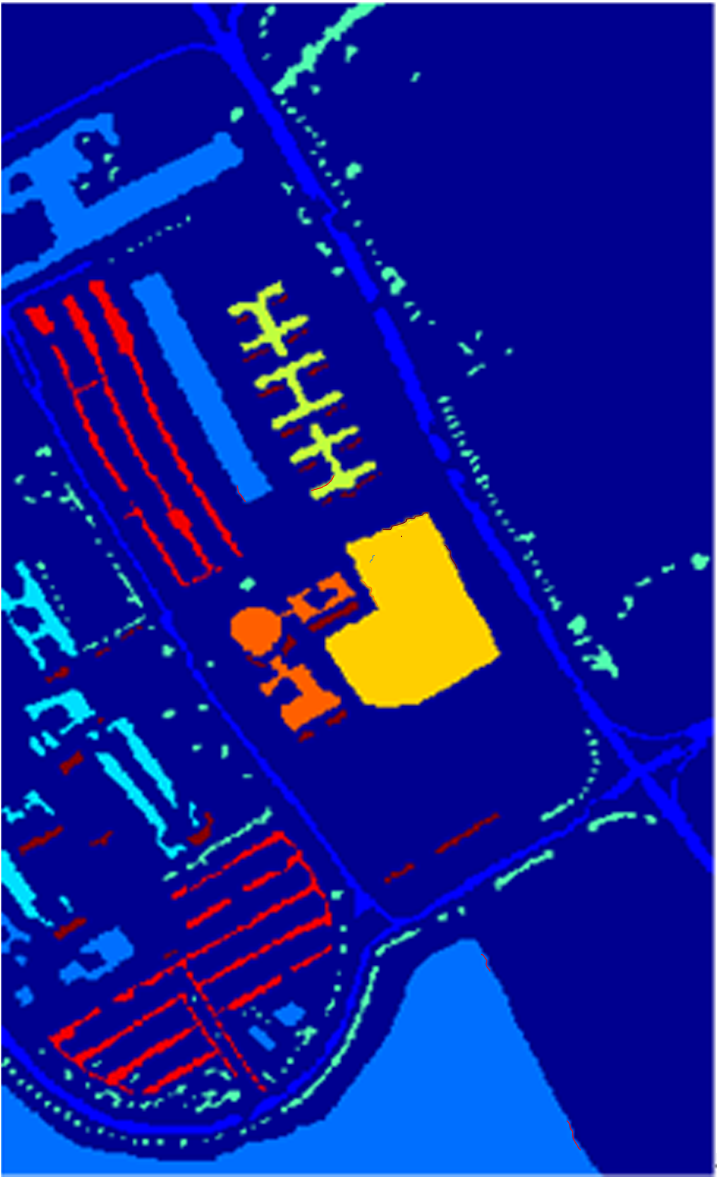}&
    \includegraphics[width=0.13\linewidth]{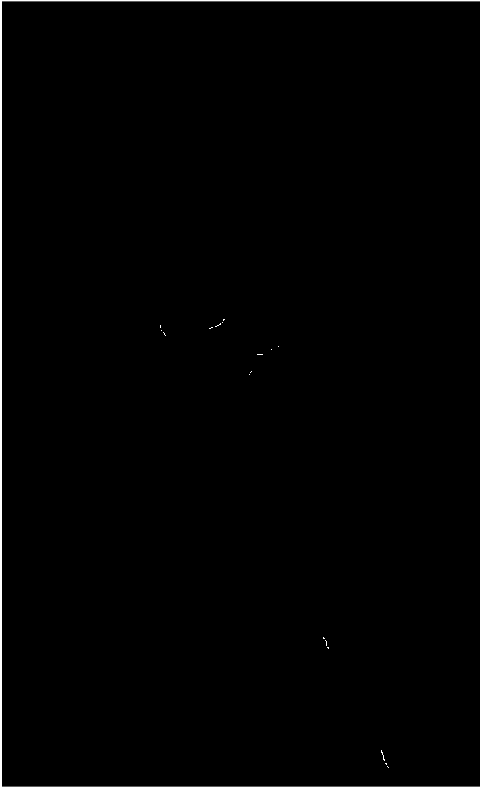}\\[2\tabcolsep]

     \includegraphics[width=0.13\linewidth]{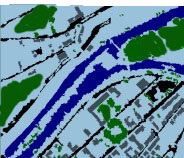}&
      \includegraphics[width=0.13\linewidth]{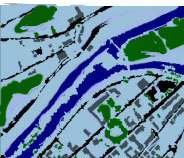}&
       \includegraphics[width=0.13\linewidth]{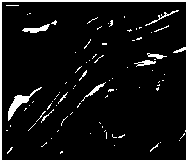}&
        \includegraphics[width=0.13\linewidth]{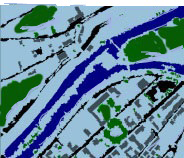}&
    \includegraphics[width=0.13\linewidth]{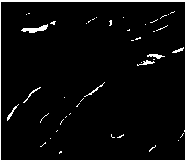}\\[2\tabcolsep]

     \includegraphics[width=0.13\linewidth]{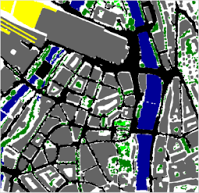}&
      \includegraphics[width=0.13\linewidth]{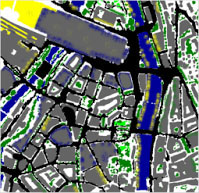}&
       \includegraphics[width=0.13\linewidth]{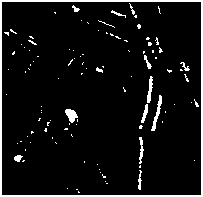}&
        \includegraphics[width=0.13\linewidth]{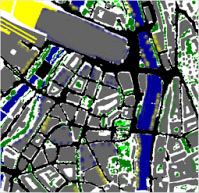}&
    \includegraphics[width=0.13\linewidth]{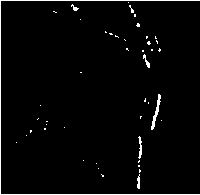}\\[2\tabcolsep]

     \includegraphics[width=0.13\linewidth]{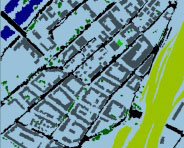}&
      \includegraphics[width=0.13\linewidth]{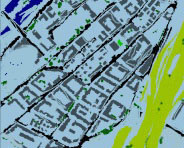}&
       \includegraphics[width=0.13\linewidth]{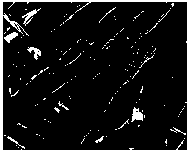}&
        \includegraphics[width=0.13\linewidth]{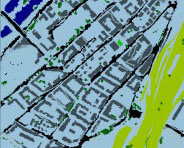}&
    \includegraphics[width=0.13\linewidth]{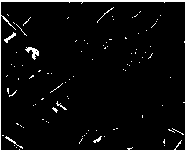}\\[2\tabcolsep]
\end{tabular}
        \caption{(a) Ground truth; (b) 3D CNN-based classification; (c) Difference map with ground truth; (d) Segmentation by deep CRF with deconvolution; and (e) Difference map after final segmentation.}
        \label{Results}
\end{figure*}

Fig.~\ref{Results} illustrates the classification and segmentation results on the Indian Pines, Pavia University and Griffith-USGS datasets respectively. The first and the second columns are the ground truth and the initial classification map on each dataset generated by 3D CNN, respectively. The third column contains binary error maps obtained by comparing the classification results with the ground truth. The white pixels indicate the parts of the image that are incorrectly classified. The fourth column shows the final segmentation outcome using deep CRF and deconvolution, whose error maps are presented in the last column. The differences between the binary maps represented in columns (c) and (e) show that the number of incorrectly segmented pixels are significantly decreased after introducing the deep CRF. This suggests the usefulness of CNN-based deep CRF with deconvolution for segmenting hyperspectral images.

\subsection{Performance Analysis and Parameter Settings}

\subsubsection{Effect of Few Spectral Bands}

\begin{figure}[h]
\centering
\includegraphics[scale=0.50]{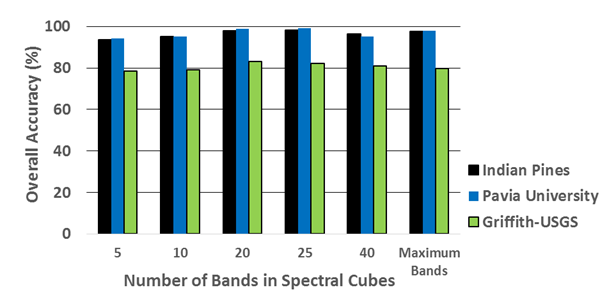}
\caption{Classification results of different numbers of bands in spectral cubes.}
\label{Optimal-bands}
\end{figure}

In our algorithm, we propose to use few spectral bands instead of the whole spectrum to construct the spectral-spatial representation of our hyperspectral data during the classification stage. We chose to use fewer bands in order to better characterize a range of spectral variability among the entire spectral signature of the data. Although during the training of CNNs, the large number of spectral cubes results in a large number of feature maps, these are able to capture the local image information well and in precise, as well as can contribute to describing the underlying materials which are very significant for segmenting hyperspectral images. Furthermore, we augmented training samples from these band groups which (1) increased the size of the training samples significantly and (2) generated more effective spectral-spatial representation of samples from different wavelengths. During the experiments, we observed that with smaller spectral cubes, we were able to detect a wider range of spectral information from our input data and hence we achieved better classification accuracy than that of using the entire spectrum. Moreover, we also analysed on the number of optimal bands to be added in individual spectral cubes by connecting this step with the data augmentation process. Since different wavelength groups capture different underlying material information, we choose the size of the band groups by testing on a various group of bands and measuring the corresponding accuracies. We discovered that the optimal number of bands should be 25 for Indian Pines and Pavia University, and 20 for Griffith-USGS. Table~\ref{SC} shows the relative comparison between these two settings of using spectral cubes for the initial classification by 3D-CNN. The analysis of choosing the optimal number of bands for respective datasets is given in Fig.~\ref{Optimal-bands} where the maximum bands denote all the bands used for representing the spectral cube.
\begin{table}[t]
\centering
\caption{Effect of Fewer Spectral Cubes}
\label{SC}
\begin{tabular}{|c|c|c|}
\hline
Dataset          & \begin{tabular}[c]{@{}c@{}}Accuracy (\%) with \\ Whole Spectral Cube\end{tabular} & \begin{tabular}[c]{@{}c@{}}Accuracy (\%) with \\ Smaller Spectral Cubes\end{tabular} \\ \hline
Indian Pines     & 96.80 $\pm$ 1.01                  & \textbf{99.20 $\pm$ 0.09                    } \\ \hline
Pavia University & 97.80 $\pm$ 1.05                  & \textbf{99.69 $\pm$ 0.03                    } \\ \hline
Griffith-USGS    & 79.65 $\pm$ 0.65                  & \textbf{84.98 $\pm$ 0.86                    } \\ \hline
\end{tabular}
\end{table}

\subsubsection{Effect of Data Augmentation}
\begin{figure}[t]
\centering
\includegraphics[scale=0.40]{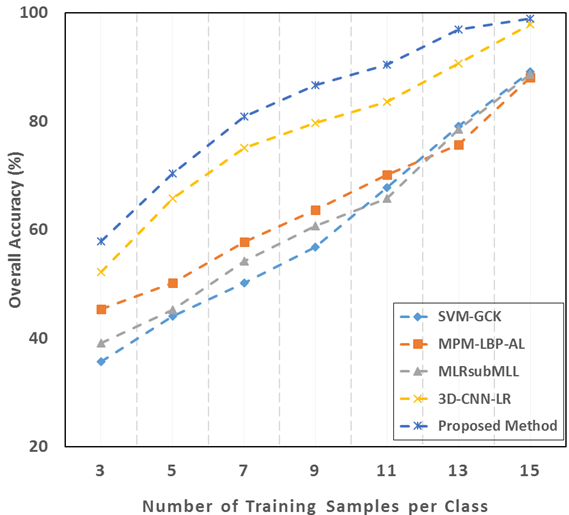}
\caption{\color{blue}Overall accuracies with different numbers of training samples per class for Indian Pines data.}
\label{sample-IP}
\end{figure}

\begin{figure}[]
\centering
\includegraphics[scale=0.40]{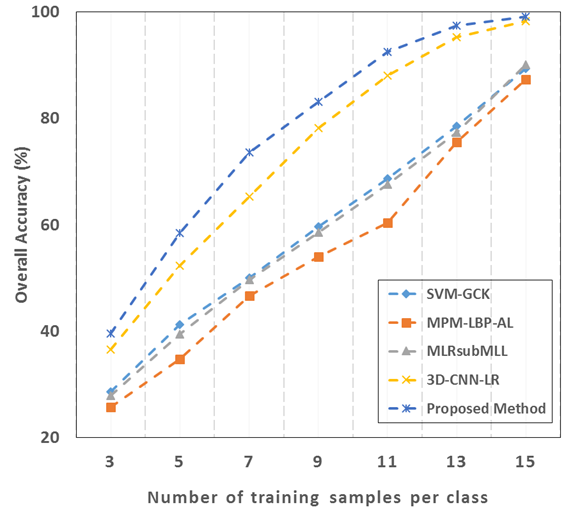}
\caption{\color{blue}Overall accuracies with different numbers of training samples per class for Pavia University data.}
\label{sample-PV}
\end{figure}

\begin{figure}[t]
\centering
\includegraphics[scale=0.40]{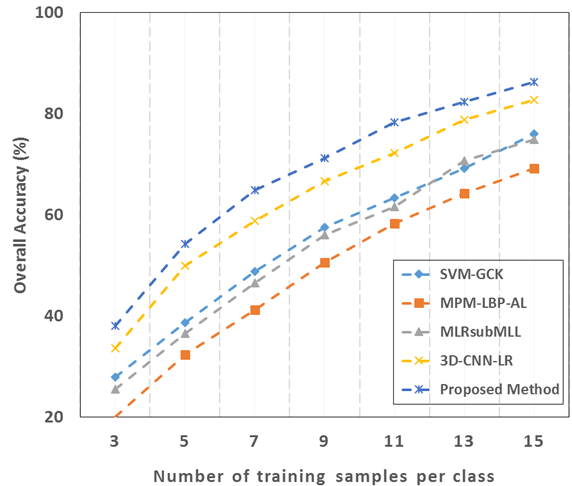}
\caption{\color{blue}Overall accuracies with different numbers of training samples per class for Griffith-USGS data.}
\label{sample-USGS}
\end{figure}

\textcolor[rgb]{0.00,0.00,1.00}{We tested our method by considering few number of training samples and then used data augmentation to avoid the problem of overfitting. During the experiments, we chose different numbers of training samples and augmented the size accordingly. We observed that increasing the numbers of training samples with augmented data obtained from different band groups had evidently improved the overall performance of our algorithm. Moreover, we also tested other methods with the same experimental settings and noticed the improved performance achieved by those as well. Figs.~\ref{sample-IP}, \ref{sample-PV} and \ref{sample-USGS} show the effect of various number of training samples which were used in data augmentation in the overall classification accuracy computed from different spectral cubes for all three datasets respectively. We also reported the overall accuracies obtained by our proposed method with and without augmenting data in Table \ref{Compare-DataAugment} for all three datasets experimented. The accuracy increased significantly by data augmentation. For instance, the accuracy improved by almost 35\% when only 3 samples were used for training the CNNs. It is quite evident from this analysis that the data augmentation had eventually contributed in improving the performance of the CNNs when limited training data are available.}

\begin{table*}[]
\centering
\caption{\color{blue}Effect of Data Augmentation on Overall Accuracies (\%) for all three datasets}
\label{Compare-DataAugment}
\begin{tabular}{|c|c|c|c|c|c|c|}
\hline
\multirow{2}{*}{\begin{tabular}[c]{@{}c@{}}\color{blue} Number of \\\color{blue} Training Samples\\\color{blue} per Class\end{tabular}}                                                                                                    & \multicolumn{2}{c|}{{\color[HTML]{3531FF} Indian Pines}}                                                                                                                             & \multicolumn{2}{c|}{{\color[HTML]{3531FF} Pavia University}}                                                                                                                         & \multicolumn{2}{c|}{{\color[HTML]{3531FF} Griffith-USGS}}                                                                                                                                 \\ \cline{2-7} 
&
{\color[HTML]{3531FF} \begin{tabular}[c]{@{}c@{}}Without\\ Data Augmentation\end{tabular}} & {\color[HTML]{3531FF} \begin{tabular}[c]{@{}c@{}}With\\ Data Augmentation\end{tabular}} & {\color[HTML]{3531FF} \begin{tabular}[c]{@{}c@{}}Without\\ Data Augmentation\end{tabular}} & {\color[HTML]{3531FF} \begin{tabular}[c]{@{}c@{}}With\\ Data Augmentation\end{tabular}} & {\color[HTML]{3531FF} \begin{tabular}[c]{@{}c@{}}Without\\ Data Augmentation\end{tabular}} & {\color[HTML]{3531FF} \begin{tabular}[c]{@{}c@{}}With\\ Data Augmentation\\\end{tabular}} \\ \hline
{\color[HTML]{3531FF} 3}                                                                                                   & {\color[HTML]{3531FF} 19.91}                                                               & {\color[HTML]{3531FF} 57.89}                                                            & {\color[HTML]{3531FF} 13.66}                                                               & {\color[HTML]{3531FF} 39.55}                                                            & {\color[HTML]{3531FF} 14.25}                                                               & {\color[HTML]{3531FF} 37.95}                                                                 \\ \hline
{\color[HTML]{3531FF} 5}                                                                                                   & {\color[HTML]{3531FF} 27.57}                                                               & {\color[HTML]{3531FF} 70.34}                                                            & {\color[HTML]{3531FF} 22.58}                                                               & {\color[HTML]{3531FF} 58.44}                                                            & {\color[HTML]{3531FF} 18.30}                                                               & {\color[HTML]{3531FF} 54.21}                                                                 \\ \hline
{\color[HTML]{3531FF} 7}                                                                                                   & {\color[HTML]{3531FF} 30.63}                                                               & {\color[HTML]{3531FF} 80.91}                                                            & {\color[HTML]{3531FF} 29.95}                                                               & {\color[HTML]{3531FF} 73.65}                                                            & {\color[HTML]{3531FF} 25.81}                                                               & {\color[HTML]{3531FF} 64.93}                                                                 \\ \hline
{\color[HTML]{3531FF} 9}                                                                                                   & {\color[HTML]{3531FF} 35.28}                                                               & {\color[HTML]{3531FF} 86.60}                                                            & {\color[HTML]{3531FF} 37.05}                                                               & {\color[HTML]{3531FF} 83.12}                                                            & {\color[HTML]{3531FF} 30.05}                                                               & {\color[HTML]{3531FF} 71.16}                                                                 \\ \hline
{\color[HTML]{3531FF} 11}                                                                                                  & {\color[HTML]{3531FF} 39.05}                                                               & {\color[HTML]{3531FF} 90.48}                                                            & {\color[HTML]{3531FF} 41.15}                                                               & {\color[HTML]{3531FF} 92.45}                                                            & {\color[HTML]{3531FF} 33.80}                                                               & {\color[HTML]{3531FF} 78.35}                                                                 \\ \hline
{\color[HTML]{3531FF} 13}                                                                                                  & {\color[HTML]{3531FF} 43.85}                                                               & {\color[HTML]{3531FF} 96.87}                                                            & {\color[HTML]{3531FF} 43.75}                                                               & {\color[HTML]{3531FF} 97.41}                                                            & {\color[HTML]{3531FF} 39.44}                                                               & {\color[HTML]{3531FF} 82.38}                                                                 \\ \hline
{\color[HTML]{3531FF} 15}                                                                                                  & {\color[HTML]{3531FF} 47.90}                                                               & {\color[HTML]{3531FF} 98.89}                                                            & {\color[HTML]{3531FF} 46.65}                                                               & {\color[HTML]{3531FF} 99.12}                                                            & {\color[HTML]{3531FF} 43.25}                                                               & {\color[HTML]{3531FF} 86.25}                                                                 \\ \hline
\end{tabular}
\end{table*}

\subsubsection{Effect of Depth in CNNs}
\begin{figure}[t]
\centering
\includegraphics[scale=0.40]{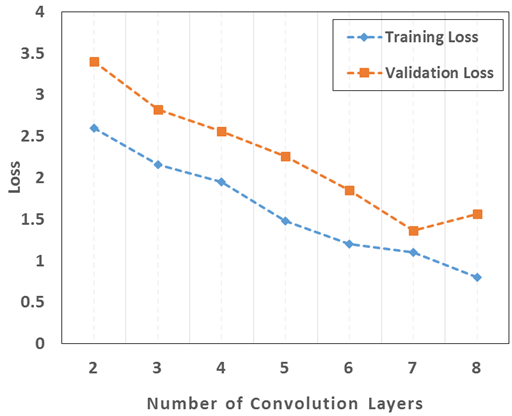}
\caption{\color{blue}Training and validation losses for Indian Pines data}
\label{depth-ip}
\end{figure}

\begin{figure}[t]
\centering
\includegraphics[scale=0.40]{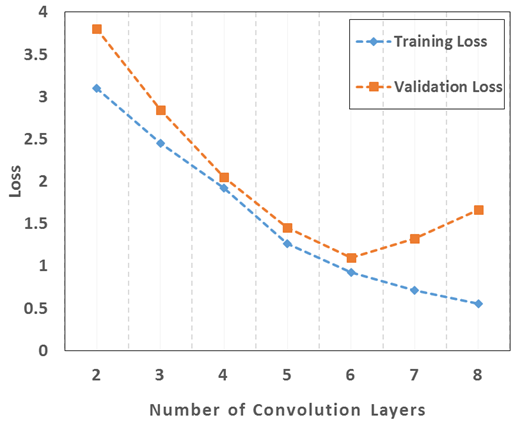}
\caption{\color{blue}Training and validation losses for Pavia University data}
\label{depth-pu}
\end{figure}

\begin{figure}[t]
\centering
\includegraphics[scale=0.40]{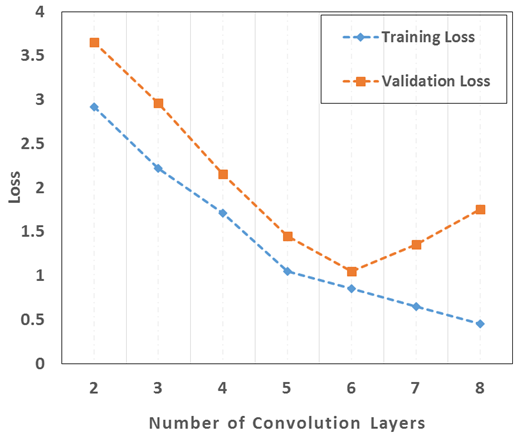}
\caption{\color{blue}Training and validation losses for Griffith-USGS data}
\label{depth-usgs}
\end{figure}

\textcolor[rgb]{0.00,0.00,1.00}{An important observation can be made from the results reported earlier in terms of the depth of the networks. Undoubtedly, depth helps in improving the classification accuracy but adding too many layers introduces the curse of overfitting and may also downgrade the accuracy as well. It is widely accepted that minimizing both training and validation losses are important in a well trained network. If the training loss is small and the validation loss is large, it means that the network is overfitted and will not generalize well for the testing samples. Therefore, we optimized the CNNs using trial and error approach and determined the number of nodes in the hidden layers, learning rate, kernel size and number of convolution layers. During the experiments, we started with a small number of convolution layers and gradually increase it. At the same time, we monitored the training and validation losses as these are varying with the changing number of layers. The effect of depth of the convolution layers for the classification stage of our algorithm is illustrated in Figs.~\ref{depth-ip},~\ref{depth-pu} and\ref{depth-usgs} for the three datasets experimented.}

\subsubsection{Influence of depth in Kernels} 
\begin{figure}[t]
\centering
\includegraphics[scale=0.40]{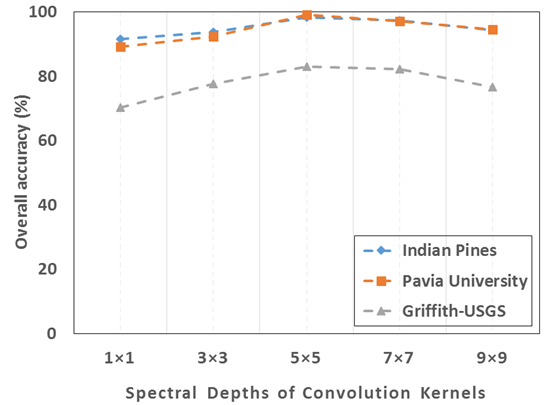}
\caption{\color{blue}Effect of spectral depth in convolution kernels}
\label{depth-kernel}
\end{figure}

The spectral depth of kernels also plays an important role in the final segmentation performance. Small receptive fields of convolution kernels generally result in better performance because in this way it is possible to learn finer details from the input. \textcolor[rgb]{0.00,0.00,1.00}{During our experiments, we varied the spectral depths of the kernels between three to nine. Fig.~\ref{depth-kernel} shows that $5\times5 \times B$ is an optimal size for all the three datasets. We found that a larger size kernel such as $9 \times 9 \times B$ ignored and skipped some essential details in the images. On the other hand, a smaller size kernel such as $3 \times 3 \times B$ provided overly detailed local information and therefore, created confusions in classification eventually. Hence, the determination of an optimal size of the kernel is important in finding the most discriminative features for classification.}

\subsubsection{Influence of ReLU} Compared to sigmoid functions, ReLU obtains better performance in terms of both complexity and accuracy~\cite{ReLU} (shown in Table \ref{ReLU}). According to our experiments, we found out that with ReLU, we achieved convergence faster than sigmoid function. For Griffith-USGS, CNN with ReLU reaches convergence almost two times faster than the same network with sigmoid. Performance was also consistently better for the other two datasets with ReLU. Furthermore, the models with ReLU can lead to lower training error at the end of training.

\begin{table}[t]
\centering
\caption{Effect of ReLU}
\label{ReLU}
\begin{tabular}{|c|c|c|c|c|}
\hline
\multirow{2}{*}{Dataset} & \multicolumn{2}{c|}{Accuracy (\%)} & \multicolumn{2}{c|}{Runtime (in minutes)} \\ \cline{2-5}
                         & Sigmoid        & ReLU         & Sigmoid               & ReLU              \\ \hline
Indian Pines             & 98.15          & \textbf{99.41}         & 57 & \textbf{36}                \\ \hline
Pavia University         & 99.04          & \textbf{99.79}        & 77 & \textbf{49}                \\ \hline
Griffith-USGS              & 85.97          & \textbf{89.13}        & 912 & \textbf{512}               \\ \hline
\end{tabular}
\end{table}

\subsection{Analysis of Computation Cost}

\textcolor[rgb]{0.00,0.07,1.00}{Here, we calculate the computational cost of segmenting an image with our trained model. The total cost combines the computational complexities for (1) generating an initial classification obtained by CNN and (2) segmenting to refine the performance by deep CRF. Generally, the convolution operations impose a significant time constraint on the time complexity of CNN which is computed in terms of the number of convolution layers, number and size of kernels and size of the intermediate feature maps~\cite{time-cnn}. The generated feature map by CNN is formulated as a CRF graph in which the voxels are represented as individual nodes. Therefore, the time complexity of CRF is computed in terms of the number of edges between the nodes as well as the size of the label set, which is quadratic in general. However, the use of highly efficient approximations for high-dimensional filtering during the message passing of mean field inference algorithm reduced the time complexity to linear in the number of labels and in the number of edges in the CRF model~\cite{mean}. Hence, the total time complexity of our algorithm is given by:}
\begin{equation}
\textcolor[rgb]{0.00,0.07,1.00}{O\left(\sum_{l=1}^D K_{l-1}.R_l^2.K_l.d_l^2\right)+O(N.Y)} 
\end{equation}
\textcolor[rgb]{0.00,0.07,1.00}{Here, $l$ is the current convolutional layer, \textit{D} is  the number of convolutional layers, $K_l$ is the number of kernels in the $l$-th layer, $K_{l-1}$ is also known as the number of input channels in the $l$-th layer, $R_l$ is the spatial size of the kernel and $d_l$ is the spatial size of the intermediate feature maps. \textit{N} is the number of edges in the CRF graph formulated from the initial CNN and \textit{Y} is the size of the label set.}

\textcolor[rgb]{0.00,0.07,1.00}{We compared the testing time for all methods included in the experiments. Since the baseline methods used in our experiments were implemented on CPU, therefore, for a fair comparison, we also chose to run our algorithm on CPU instead of GPU that is widely used for deep learning approaches. All methods were implemented in Matlab and few C modules, and run on a desktop computer with Intel Core i5-4570 @ 3.2GHz 8G memory, with a Windows 7 system.}

\textcolor[rgb]{0.00,0.07,1.00}{The results are shown in Table~\ref{runtime-total}. The testing stage of the deep learning algorithms are very fast and are close to the time required by other baseline methods. This is an important property for real applications as the model training can be undertaken offline but the application of the trained model on new data has higher efficiency requirements.}

\begin{table}[]
\centering
\caption{\color[HTML]{3531FF} RUNNING TIME COMPARISON (MEASURED IN MINUTES)}
\label{runtime-total}
\begin{tabular}{|c|c|c|}
\hline
{\color[HTML]{3531FF} Methods}                                   & {\color[HTML]{3531FF} Dataset}          & {\color[HTML]{3531FF} Testing Time} \\ \hline
{\color[HTML]{3531FF} }                                          & {\color[HTML]{3531FF} Indian Pines}     & {\color[HTML]{3531FF} 0.63}      \\ \cline{2-3} 
{\color[HTML]{3531FF} }                                          & {\color[HTML]{3531FF} Pavia University} & {\color[HTML]{3531FF} 0.91}      \\ \cline{2-3} 
\multirow{-3}{*}{{\color[HTML]{3531FF} MPM-LBP-AL~\cite{spatial}}} & {\color[HTML]{3531FF} Griffith-USGS}    & {\color[HTML]{3531FF} 0.93}      \\ \hline
{\color[HTML]{3531FF} }                                          & {\color[HTML]{3531FF} Indian Pines}     & {\color[HTML]{3531FF} 0.58}      \\ \cline{2-3} 
{\color[HTML]{3531FF} }                                          & {\color[HTML]{3531FF} Pavia University} & {\color[HTML]{3531FF} 0.88}      \\ \cline{2-3} 
\multirow{-3}{*}{{\color[HTML]{3531FF} MLRsubMLL~\cite{Li}}}       & {\color[HTML]{3531FF} Griffith-USGS}    & {\color[HTML]{3531FF} 0.91}      \\ \hline
{\color[HTML]{3531FF} }                                          & {\color[HTML]{3531FF} Indian Pines}     & {\color[HTML]{3531FF} 0.77}      \\ \cline{2-3} 
{\color[HTML]{3531FF} }                                          & {\color[HTML]{3531FF} Pavia University} & {\color[HTML]{3531FF} 1.14}      \\ \cline{2-3} 
\multirow{-3}{*}{{\color[HTML]{3531FF} WHED~\cite{Watershed}}}     & {\color[HTML]{3531FF} Griffith-USGS}    & {\color[HTML]{3531FF} 1.15}      \\ \hline
{\color[HTML]{3531FF} }                                          & {\color[HTML]{3531FF} Indian Pines}     & {\color[HTML]{3531FF} 0.91}      \\ \cline{2-3} 
{\color[HTML]{3531FF} }                                          & {\color[HTML]{3531FF} Pavia University} & {\color[HTML]{3531FF} 1.12}      \\ \cline{2-3} 
\multirow{-3}{*}{{\color[HTML]{3531FF} 3D-CNN-LR~\cite{yushi}}}    & {\color[HTML]{3531FF} Griffith-USGS}    & {\color[HTML]{3531FF} 1.09}      \\ \hline
{\color[HTML]{3531FF} }                                          & {\color[HTML]{3531FF} Indian Pines}     & {\color[HTML]{3531FF} 0.79}      \\ \cline{2-3} 
{\color[HTML]{3531FF} }                                          & {\color[HTML]{3531FF} Pavia University} & {\color[HTML]{3531FF} 1.05}      \\ \cline{2-3} 
\multirow{-3}{*}{{\color[HTML]{3531FF} Proposed Method}}         & {\color[HTML]{3531FF} Griffith-USGS}    & {\color[HTML]{3531FF} 1.08}      \\ \hline
\end{tabular}
\end{table}

\section{Conclusion}

We presented an efficient CRF-CNN based deep learning algorithm for segmenting hyperspectral images. To utilize the full strength of deep models for complex computer vision tasks, we constructed a powerful spatial-spectral representation of hyperspectral data. We applied 3D-CNN in a range of more effective spectral-spatial representative band groups to extract initial features. To further facilitate the segmentation task, we integrated CRF with CNN into a common framework in which the parameters of CRF are calculated using CNN, therefore making it a deep CRF. The initial prediction results coming from this CRF-CNN architecture was further improved by using a deconvolution block inside of the CRF pairwise potential calculations. Moreover, to overcome the problem of over-fitting, we employed data augmentation techniques and increased the size of training samples for training the CNNs. This effectively improved the overall performance of our deep network to a significant extent.

In summary, to achieve the improvement of the hyperspectral image segmentation performance, we proposed an architecture containing several important efficient stages that not only optimized the calculations of such computationally expensive task but also improved the initial prediction results obtained by the initial CNN algorithm. With this method, we can fully exploit the usefulness of CRF in the context of segmentation by integrating it completely inside of a deep learning algorithm. We further evaluated the usefulness of our method by comparing it with several state-of-the-art methods and achieved promising results.

\bibliographystyle{IEEEbib}
\bibliography{strings,refs}

\begin{thebibliography}{10}

\bibitem{HSI}
D.~Landgrebe,
\newblock ``Hyperspectral image data analysis,''
\newblock {\em IEEE Signal Processing Magazine}, vol. 19, no. 1, pp. 17--28,
  2002.

\bibitem{Remote}
J.~Richards,
\newblock {\em Remote Sensing Digital Image Analysis: An Introduction},
\newblock Springer Berlin Heidelberg, NJ, USA, 5th edition, 2013.

\bibitem{kernel}
G.~Camps-Valls and L.~Bruzzone,
\newblock ``Kernel-based methods for hyperspectral image classification,''
\newblock {\em IEEE Transactions on Geoscience and Remote Sensing}, vol. 43,
  no. 6, pp. 1351--1362, 2005.

\bibitem{Spec-Spatial}
M.~Fauvel, Y.~Tarabalka, J.~Benediktsson, J.~Chanussot, and J.~Tilton,
\newblock ``Advances in spectral-spatial classification of hyperspectral
  images,''
\newblock {\em Proceedings of the IEEE}, vol. 101, no. 3, pp. 652--675, 2013.

\bibitem{attribute}
P.~Ghamisi, M.~Mura, and J.~Benediktsson,
\newblock ``A survey on spectral-spatial classification techniques based on
  attribute profiles,''
\newblock {\em IEEE Transactions on Geoscience and Remote Sensing}, vol. 53,
  no. 5, pp. 2335--2353, 2015.

\bibitem{meanshift}
C.~Deng, S.~Li, F.~Bian, and Y.~Yang,
\newblock {\em Remote Sensing Image Segmentation Based on Mean Shift Algorithm
  with Adaptive Bandwidth}, pp. 179--185,
\newblock Springer Berlin Heidelberg, Berlin, Heidelberg, 2015.

\bibitem{SVM-MRF}
Y.~Tarabalka, M.~Fauvel, J.~Chanussot, and J.~Benediktsson,
\newblock ``{SVM} and {MRF}-based method for accurate classification of
  hyperspectral images,''
\newblock {\em IEEE Geoscience and Remote Sensing Letters}, vol. 7, no. 4, pp.
  736--740, 2010.

\bibitem{MRF2}
Y.~Zhao, L.~Zhang, P.~Li, and B.~Huang,
\newblock ``Classification of high spatial resolution imagery using improved
  {Gaussian Markov} random-field-based texture features,''
\newblock {\em IEEE Transactions on Geoscience and Remote Sensing}, vol. 45,
  no. 5, pp. 1458--1468, 2007.

\bibitem{MRF-seg}
O.~Eches, J.~Benediktsson, N.~Dobigeon, and J.~Tourneret,
\newblock ``Adaptive {Markov} random fields for joint unmixing and segmentation
  of hyperspectral images,''
\newblock {\em IEEE Transactions on Image Processing}, vol. 22, no. 1, pp.
  5--16, 2013.

\bibitem{Li}
J.~Li, J.~Bioucas-Dias, and A.~Plaza,
\newblock ``Spectral-spatial hyperspectral image segmentation using subspace
  multinomial logistic regression and {Markov} random fields,''
\newblock {\em IEEE Transactions on Geoscience and Remote Sensing}, vol. 50,
  no. 3, pp. 809--823, 2012.

\bibitem{ML}
J.~Li, J.~Bioucas-Dias, and A.~Plaza,
\newblock ``Semisupervised hyperspectral image segmentation using multinomial
  logistic regression with active learning,''
\newblock {\em IEEE Transactions on Geoscience and Remote Sensing}, vol. 48,
  no. 11, pp. 4085--4098, 2010.

\bibitem{spectraltexture}
J.~Yuan, D.~Wang, and R.~Li,
\newblock ``Remote sensing image segmentation by combining spectral and texture
  features,''
\newblock {\em IEEE Transactions on Geoscience and Remote Sensing}, vol. 52,
  no. 1, pp. 16--24, 2014.

\bibitem{graph}
D.~Gills and J.~Bowles,
\newblock ``Hyperspectral image segmentation using spatial-spectral graphs,''
\newblock in {\em Algorithms and Technologies for Multispectral, Hyperspectral,
  and Ultraspectral Imagery XVIII}, 2012, vol. 8390, pp. 83901Q--1--83901Q--11.

\bibitem{Watershed}
Y.~Tarabalka, J.~Chanussot, and J.~Benediktsson,
\newblock ``Segmentation and classification of hyperspectral images using
  watershed transformation,''
\newblock {\em Pattern Recognition}, vol. 43, no. 7, pp. 2367--2379, 2010.

\bibitem{MSF}
K.~Bernard, Y.~Tarabalka, J.~Angulo, J.~Chanussot, and J.~Benediktsson,
\newblock ``Spectral-spatial classification of hyperspectral data based on a
  stochastic minimum spanning forest approach,''
\newblock {\em IEEE Transactions on Image Processing}, vol. 21, no. 4, pp.
  2008--2021, 2012.

\bibitem{DL}
N.~Kruger, P.~Janssen, S.~Kalkan, M.~Lappe, A.~Leonardis, J.~Piater,
  A.~Rodriguez-Sanchez, and L.~Wiskott,
\newblock ``Deep hierarchies in the primate visual cortex: What can we learn
  for computer vision?,''
\newblock {\em IEEE Transactions on Pattern Analysis and Machine Intelligence},
  vol. 35, no. 8, pp. 1847--1871, 2013.

\bibitem{Jia}
Y.~Chen, X.~Zhao, and X.~Jia,
\newblock ``Spectral-spatial classification of hyperspectral data based on deep
  belief network,''
\newblock {\em IEEE Journal of Selected Topics in Applied Earth Observations
  and Remote Sensing}, vol. 8, no. 6, pp. 2381--2392, 2015.

\bibitem{DeepCNN}
H.~Lee and H.~Kwon,
\newblock ``Going deeper with contextual {CNN} for hyperspectral image
  classification,''
\newblock {\em IEEE Transactions on Image Processing}, vol. 26, no. 10, pp.
  4843--4855, 2017.

\bibitem{FCN}
E.~Shelhamer, J.~Long, and T.~Darrell,
\newblock ``Fully convolutional networks for semantic segmentation,''
\newblock in {\em Proceedings of the IEEE Conference on Computer Vision and
  Pattern Recognition}, 2015, pp. 3431--3440.

\bibitem{yushi}
Y.~Chen, H.~Jiang, C.~Li, X.~Jia, and P.~Ghamisi,
\newblock ``Deep feature extraction and classification of hyperspectral images
  based on convolutional neural networks,''
\newblock {\em IEEE Transactions on Geoscience and Remote Sensing}, vol. 54,
  no. 10, pp. 6232--6251, 2016.

\bibitem{shiqi}
Shiqi Yu, Sen Jia, and Chunyan Xu,
\newblock ``Convolutional neural networks for hyperspectral image
  classification,''
\newblock {\em Neurocomputing}, vol. 219, pp. 88 -- 98, 2017.

\bibitem{blde-cnn}
W.~Zhao and S.~Du,
\newblock ``Spectral-spatial feature extraction for hyperspectral image
  classification: A dimension reduction and deep learning approach,''
\newblock {\em IEEE Transactions on Geoscience and Remote Sensing}, vol. 54,
  no. 8, pp. 4544--4554, 2016.

\bibitem{pixel-pair}
W.~Li, G.~Wu, F.~Zhang, and Q.~Du,
\newblock ``Hyperspectral image classification using deep pixel-pair
  features,''
\newblock {\em IEEE Transactions on Geoscience and Remote Sensing}, vol. 55,
  no. 2, pp. 844--853, 2017.

\bibitem{CNN-prob}
S.~C. Douglas,
\newblock ``A novel endmember, fractional abundance, and contrast model for
  hyperspectral imagery,''
\newblock in {\em 2013 IEEE International Conference on Acoustics, Speech and
  Signal Processing}, 2013, pp. 2164--2168.

\bibitem{CNN-CRF2}
A.~Kirillov, D.~Schlesinger, S.~Zheng, B.~Savchynskyy, P.~Torr, and C.~Rother,
\newblock ``Joint training of generic {CNN-CRF} models with stochastic
  optimization,''
\newblock in {\em 13th Asian Conference on Computer Vision}, 2016, pp.
  221--236.

\bibitem{CNN-CRF3}
X.~Chu, W.~Ouyang, H.~Li, and X.~Wang,
\newblock ``Structured feature learning for pose estimation,''
\newblock in {\em IEEE Conference on Computer Vision and Pattern Recognition},
  2016, pp. 4715--4723.

\bibitem{CNN-CRF4}
X.~Chu, W.~Ouyang, H.~Li, and X.~Wang,
\newblock ``{CRF-CNN}: Modeling structured information in human pose
  estimation,''
\newblock in {\em Advances in Neural Information Processing Systems}, 2016, pp.
  316--324.

\bibitem{CNN-CRF5}
P.~Kn{\"{o}}belreiter, C.~Reinbacher, A.~Shekhovtsov, and T.~Pock,
\newblock ``End-to-end training of hybrid {CNN-CRF} models for stereo,''
\newblock in {\em Proceedings of the IEEE Conference on Computer Vision and
  Pattern Recognition}, 2017, pp. 2339--2348.

\bibitem{CRF-CNN}
F.~Liu, G.~Lin, and C.~Shen,
\newblock ``{CRF} learning with {CNN} features for image segmentation,''
\newblock {\em Pattern Recognition}, vol. 48, no. 10, pp. 2983--2992, 2015.

\bibitem{deconvo}
L.~Chen, G.~Papandreou, I.~Kokkinos, K.~Murphy, and A.~Yuille,
\newblock ``Semantic image segmentation with deep convolutional nets and fully
  connected {CRF}s,''
\newblock in {\em Proceedings of the International Conference on Learning
  Representations}, 2015.

\bibitem{CRF-RNN}
S.~Zheng, S.~Jayasumana, B.~Romera-Paredes, V.~Vineet, Z.~Su, D.~Du, C.~Huang,
  and P.~Torr,
\newblock ``Conditional random fields as recurrent neural networks,''
\newblock in {\em Proceedings of the 2015 IEEE International Conference on
  Computer Vision}, 2015, pp. 1529--1537.

\bibitem{deconvo3}
M.~Zeiler, G.~Taylor, and R.~Fergus,
\newblock ``Adaptive deconvolutional networks for mid and high level feature
  learning,''
\newblock in {\em Proceedings of the 2011 International Conference on Computer
  Vision}, 2011, pp. 2018--2025.

\bibitem{piecewise2}
G.~Lin, C.~Shen, A.~Hengel, and I.~Reid,
\newblock ``Efficient piecewise training of deep structured models for semantic
  segmentation,''
\newblock in {\em Proceedings of the IEEE Conference on Computer Vision and
  Pattern Recognition}, 2016, pp. 3194--3203.

\bibitem{mean}
P.~Kr\"{a}henb\"{u}hl and V.~Koltun,
\newblock ``Efficient inference in fully connected {CRFs} with gaussian edge
  potentials,''
\newblock in {\em Advances in Neural Information Processing Systems}, pp.
  109--117. 2011.

\bibitem{Autoencoder}
L.~Zhang, L.~Zhang, and B.~Du,
\newblock ``Deep learning for remote sensing data: A technical tutorial on the
  state of the art,''
\newblock {\em IEEE Geoscience and Remote Sensing Magazine}, vol. 4, no. 2, pp.
  22--40, 2016.

\bibitem{3D-CNN-2}
Y.~Li, H.~Zhang, and Q.~Shen,
\newblock ``Spectral spatial classification of hyperspectral imagery with {3D}
  convolutional neural network,''
\newblock {\em Remote Sensing}, vol. 9, no. 1, 2017.

\bibitem{CRF-train}
A.~Kolesnikov, M.~Guillaumin, V.~Ferrari, and C.~H. Lampert,
\newblock ``Closed-form training of conditional random fields for large scale
  image segmentation,''
\newblock in {\em Proceedings of the 13th European Conference on Computer
  Vision)}, 2014, pp. 550--565.

\bibitem{piecewise}
C.~Sutton and A.~McCallum,
\newblock ``Piecewise training for undirected models,''
\newblock in {\em Proceedings of the Conference on Uncertainty in Artificial
  Intelligence}, 2005, pp. 568--575.

\bibitem{FLAASH}
``Fast line-of-sight atmospheric analysis of hypercubes ({FLAASH}),''
  https://www.harrisgeospatial.com/docs/FLAASH.html.

\bibitem{seg-class}
P.~Kaiser, J.~Wegner, A.~Lucchi, M.~Jaggi, T.~Hofmann, and K.~Schindler,
\newblock ``Learning aerial image segmentation from online maps,''
\newblock {\em IEEE Transactions on Geoscience and Remote Sensing}, vol. 55,
  no. 11, pp. 6054--6068, 2017.

\bibitem{spatial}
J.~Li, J.~Bioucas-Dias, and A.~Plaza,
\newblock ``Spectral-spatial classification of hyperspectral data using loopy
  belief propagation and active learning,''
\newblock {\em IEEE Transactions on Geoscience and Remote Sensing}, vol. 51,
  no. 2, pp. 844--856, 2013.

\bibitem{composite}
J.~Li, P.~Marpu, A.~Plaza, J.~Bioucas{-}Dias, and J.~Benediktsson,
\newblock ``Generalized composite kernel framework for hyperspectral image
  classification,''
\newblock {\em IEEE Transactions on Geoscience and Remote Sensing}, vol. 51,
  no. 9, pp. 4816--4829, 2013.

\bibitem{ReLU}
A.~Krizhevsky, I.~Sutskever, and G.~Hinton,
\newblock ``Imagenet classification with deep convolutional neural networks,''
\newblock in {\em Proceedings of the 25th International Conference on Neural
  Information Processing Systems}, 2012, pp. 1097--1105.

\bibitem{time-cnn}
K.~He and J.~Sun,
\newblock ``Convolutional neural networks at constrained time cost,''
\newblock in {\em 2015 IEEE Conference on Computer Vision and Pattern
  Recognition (CVPR)}, 2015, pp. 5353--5360.

\end{thebibliography}

\end{document}